\documentclass[lettersize,journal]{IEEEtran}
\usepackage{amsmath,amsfonts}
\usepackage{algorithmic}
\usepackage{algorithm}
\usepackage{array}
\usepackage{textcomp}
\usepackage{stfloats}
\usepackage{url}
\usepackage{verbatim}
\usepackage{graphicx}
\usepackage{cite}
\usepackage{booktabs}
\usepackage{adjustbox}
\usepackage{bm}
\usepackage{subfigure}
\usepackage{amssymb}
\usepackage[pagebackref,breaklinks,colorlinks]{hyperref}
\newcommand{\tmmacc}[2]{#1\,{\scriptsize$\pm$\,#2}}
\newcommand{\tmmbest}[2]{\textbf{#1}\,{\scriptsize$\pm$\,#2}}
\newcommand{\tmmsecond}[2]{\underline{#1}\,{\scriptsize$\pm$\,#2}}

\begin{document}

\title{DLER: Dual-Logit Experience Replay for Online Continual Learning}

\author{Huiwei Lin, Shanshan Feng, and Kong Aik Lee, Senior Member, IEEE

\thanks{
    Huiwei Lin and Kong Aik Lee are with The Hong Kong Polytechnic University, Hong Kong SAR, China (e-mail: felixlinhuiwei@gmail.com, kong-aik.lee@polyu.edu.hk).
    
    Shanshan Feng is with Wuhan University, Wuhan 430072, China (e-mail: victor\_fengss@whu.edu.cn).

}
\thanks{Manuscript received April 19, 2021; revised August 16, 2021.}
}

\markboth{Journal of \LaTeX\ Class Files,~Vol.~14, No.~8, August~2021}%
{Shell \MakeLowercase{\textit{et al.}}: A Sample Article Using IEEEtran.cls for IEEE Journals}

\IEEEpubid{0000--0000/00\$00.00~\copyright~2021 IEEE}

\maketitle

\begin{abstract}
Online continual learning aims to continuously train neural networks on a non-stationary data stream in a single pass. Rehearsal-based methods alleviate catastrophic forgetting by replaying stored samples, but the imbalance between current and previous data often leads to predictions biased toward current classes. Many approaches have been proposed to overcome the forgetting problem by correcting the bias; however, they still need to be improved in an online fashion. By decomposing the dot-product logits into an angle factor and a norm factor, our analysis reveals that the angle factor efficiently captures novel knowledge but is particularly susceptible to prediction bias, whereas the norm factor helps preserve historical knowledge. Based on these observations, we propose Dual-Logit Experience Replay (DLER), which exploits dual-logits for both model optimization and memory management. Specifically, DLER learns current samples using cosine logits and incorporates additional dot-product logits when replaying previous samples to mitigate prediction bias. To strengthen dual-logit learning under limited replay memory, an auxiliary supervised contrastive objective promotes angular compactness within each class, extending corrections learned from limited replay samples to other samples of the same class. Meanwhile, we introduce a memory update strategy that preferentially retains samples for which dot-product prediction assigns a higher probability to the correct class than cosine prediction, aligning sample selection with dual-logit training. Extensive experiments on three datasets demonstrate the effectiveness of DLER, particularly under limited memory budgets.
\end{abstract}

\section{Introduction}
\label{sec:introduction}

At present, deep neural networks~\cite{nie2022search,nie2019multimodal,nie2017enhancing} still lack the conventional ability of continuous learning like human beings. If the model adapts to new data through fine-tuning, it will significantly degrade the performance on old data. Such a phenomenon is named catastrophic forgetting (CF)~\cite{delange2021continual}. Because solving this problem is highly significant, continual learning~\cite{liu2021lifelong,liu2022ai,liu2020lifelong} has been proposed to train models to accumulate knowledge without forgetting. In particular, online continual learning (OCL) considers a practical setting in which models must mitigate CF while learning continuously from a potentially infinite, non-stationary data stream, with each incoming sample accessible only once.

Rehearsal-based methods have demonstrated strong performance in OCL~\cite{lin2022anchor}. These methods maintain a fixed-size episodic memory buffer to store previous samples and replay them while learning from current data~\cite{yin2021mitigating,mai2022online,lin2023pcr}. However, the imbalance between current data and limited previous samples can bias predictions toward new classes, degrading performance on previously learned classes. Although existing approaches mitigate forgetting by correcting this bias, three challenges remain in online settings:
1) \textbf{Real-time ability}. Some methods~\cite{wu2019large,zhao2020maintaining,mai2021supervised} require additional training steps or post-processing, hindering real-time prediction. 2) \textbf{Flexibility}. Some methods~\cite{ahn2021ss,caccia2021new} rely heavily on explicit boundaries between old and new classes, and their correction mechanisms cannot be readily extended from the predictor to the feature extractor. 3) \textbf{Adjustability}. Other methods~\cite{chrysakisonline,caccia2021new} offer limited control over the strength of bias correction, potentially restricting model generalization. Hence, can a straightforward and efficient approach address prediction bias in OCL while simultaneously meeting these three challenges?

\begin{figure*}[t]
\centering
\includegraphics[scale=0.70]{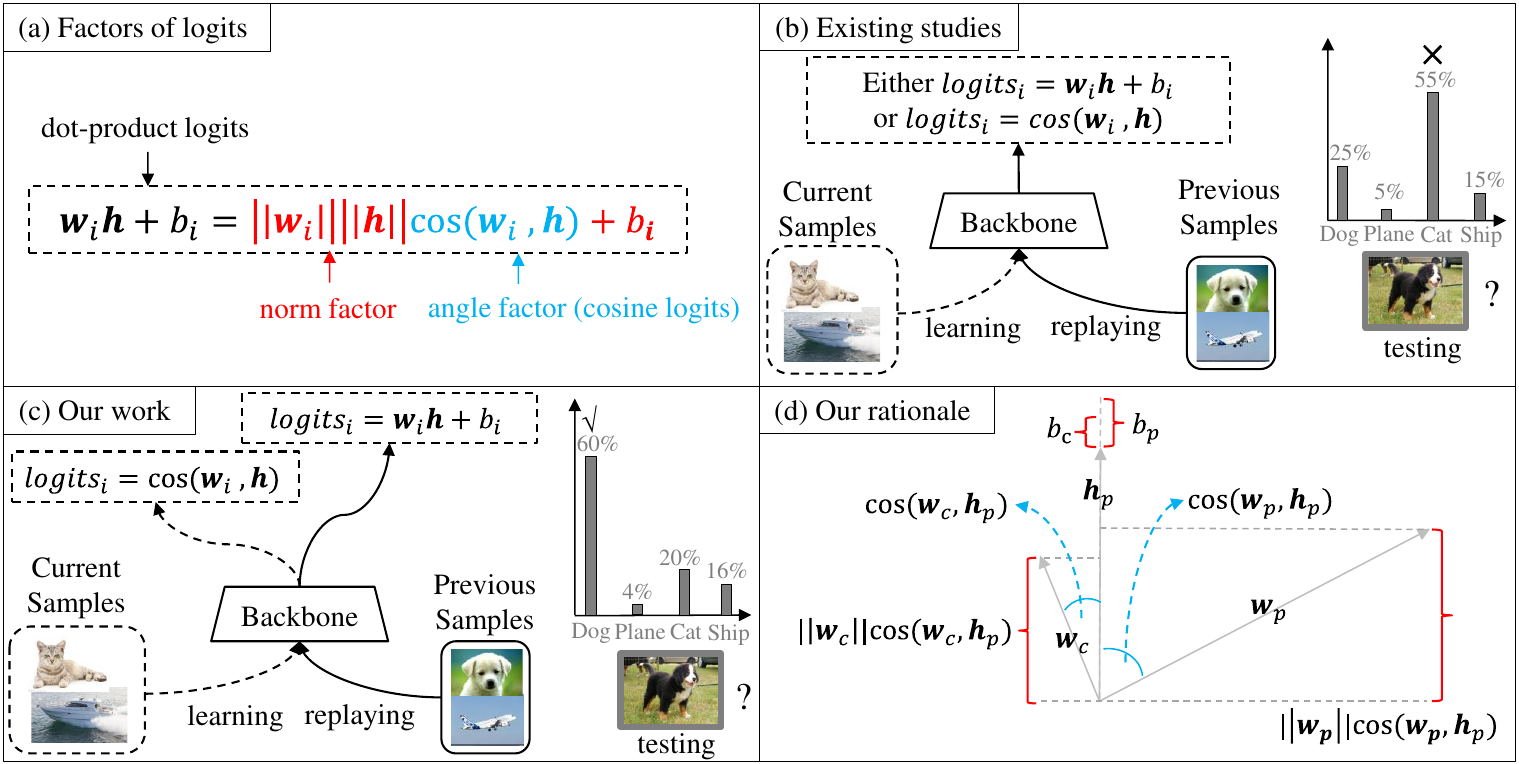}
\vspace{-2ex}
\caption{Illustration of our work. (a) The dot-product logits can be decomposed into a norm factor and an angle factor; (b) Existing studies, which learn all samples by calculating logits in the same way; (c) Different from existing studies, the model in our work learns current samples only by the angle factor, while replays previous samples by both the angle factor and the norm factor. (d) By leveraging the norm factor, the model can correctly classify samples.}
\label{fig:problem}%
\vspace{-2ex}
\end{figure*}

To answer this question, we examine existing rehearsal-based methods and empirically find that the biased predictions (dot-product logits) mainly change in two parts, as shown in Figure~\ref{fig:problem}(a). Inspired by~\cite{liu2018decoupled}, we express dot-product logits as $\bm{w}\bm{h}+b=\Vert\bm{w}\Vert\Vert\bm{h}\Vert\cos(\bm{w},\bm{h})+b$, separating the norm and angle factors while retaining the additive bias term. Data imbalance leads to unbalanced competition during gradient propagation, contributing to CF and affecting both factors. Further experiments reveal that these two factors act differently during the training process. On the one hand, the angle factor suffers from substantial prediction bias but efficiently captures novel knowledge from current samples when used independently as cosine logits. On the other hand, the norm factor, which is abandoned by existing works~\cite{hou2019learning,caccia2021new}, is helpful in keeping historical knowledge of previous samples.

\IEEEpubidadjcol

These observations suggest that it is feasible to leverage the norm factor to balance old and new knowledge and address the bias issue. Specifically, dot-product logits incorporate both norm and angular information, making them suitable for consolidating historical knowledge during replay. In contrast, cosine logits rely on angular information and facilitate the acquisition of new knowledge from current samples. Different from existing studies (Figure~\ref{fig:problem}(b)) that calculate logits in the same way, we try to (Figure~\ref{fig:problem}(c)) learn current samples by cosine logits and further replay previous samples by dot-product logits. As illustrated in Figure~\ref{fig:problem}(d), a sample $\bm{h}_p$ will be mistakenly classified to current class $c$ by cosine logits that satisfies $cos(\bm{w}_c,\bm{h}_p)>cos(\bm{w}_p,\bm{h}_p)$. However, with the help of the norm factor, the model can correctly classify $\bm{h}_p$ to the previous class $p$ by dot-product logits that satisfy $\Vert\bm{h}_p\Vert\Vert\bm{w}_c\Vert cos(\bm{w}_c,\bm{h}_p)+b_c<\Vert\bm{h}_p\Vert\Vert\bm{w}_p\Vert cos(\bm{w}_p,\bm{h}_p)+b_p$. Empirical experiments demonstrate that this approach effectively solves the bias problem and achieves better performance.

Motivated by these observations, we propose Dual-Logit Experience Replay (DLER), which alleviates CF in OCL by exploiting the complementary properties of cosine and dot-product logits to address prediction bias. Its learning process comprises three components: 1) The \textbf{learning component} trains current samples using cosine logits to acquire novel knowledge through angular information. 2) The \textbf{replaying component} incorporates additional dot-product supervision for buffered samples, leveraging norm and angular information to consolidate historical knowledge. A weighting parameter controls the contribution of dot-product supervision, allowing adjustment of the balance between current learning and historical replay. 3) The \textbf{testing component} uses dot-product logits to predict over all observed classes, retaining the contribution of norm information without introducing additional modules.

To further exploit the complementary roles of dual-logits under limited replay memory, we strengthen dual-logit training and extend it to memory management. Specifically, an auxiliary supervised contrastive loss promotes angular compactness within classes and separation between classes, providing more consistent angular evidence to complement norm information in dot-product logits. This geometry can help corrections learned through dot-product supervision generalize from limited replay samples to unseen samples of the same historical class. Moreover, we align memory selection with dual-logit training by scoring samples using the latest model according to the positive probability advantage of dot-product prediction over cosine prediction for the correct class. Within a fixed memory budget, class quotas regulate allocation, while score-based replacement preferentially retains samples exhibiting this advantage. This helps memory better support dual-logit complementarity when few samples can be retained per class. Together, these extensions exploit dual logits in model optimization and memory management, while retaining direct inference (\textbf{Real-time ability}), applicability to the feature extractor (\textbf{Flexibility}), and controllable bias correction (\textbf{Adjustability}).

The main contributions are summarized as follows:
\begin{itemize}
    \item We theoretically analyze the characteristics of dot-product logits and cosine logits, discovering that their coupling is beneficial. To the best of our knowledge, this work is the first to combine these two logits for the OCL.
    
    \item We propose DLER, a rehearsal-based framework that learns current samples using cosine logits and incorporates additional dot-product supervision when replaying previous samples.
    
    \item We further enhance dual-logit learning through an auxiliary supervised contrastive loss that promotes feature compactness and a memory update strategy that exploits the predictive complementarity of both logits.

    \item We conduct extensive experiments on Split CIFAR100, Split MiniImageNet, and Split TinyImageNet to evaluate the effectiveness of DLER. The code is publicly available at \url{https://github.com/FelixHuiweiLin/UER}.
\end{itemize}

This work extends our conference version presented in~\cite{lin2023uer} in three aspects: (i) We introduce an auxiliary supervised contrastive objective to promote angular compactness and class separation, helping dual-logit corrections generalize beyond limited replay samples. (ii) We extend dual-logits to memory management through advantage-based scoring and replacement under balanced class quotas, prioritizing samples with a positive dot-product probability advantage for the correct class. (iii) We replace Split CIFAR10 with the more challenging Split TinyImageNet and conduct additional ablation studies and analyses to examine the effectiveness and mechanisms of the proposed components.

\section{Related Work}
\label{sec:relatedwork}

\subsection{Continual Learning}

Continual learning, also referred to as lifelong or incremental learning, enables models to acquire knowledge from sequentially arriving data while preserving previously learned knowledge. Its central goal is to balance adaptation to new data with resistance to catastrophic forgetting. Existing approaches can be categorized into three groups. 
1) Architecture-based methods~\cite{savadikar2026cheem} mitigate CF through dynamic or static network designs. Dynamic networks progressively expand their capacity as learning proceeds, whereas static networks retain a fixed architecture and selectively allocate parameters. 2) Regularization-based methods~\cite{liu2026ewcdr} introduce additional constraints into the training objective to preserve historical knowledge, typically by penalizing changes to important parameters or applying knowledge distillation. 3) Rehearsal-based methods~\cite{deng2025rehearsal,cui2021deepcollaboration} retain a subset of historical samples or generate synthetic samples, which are revisited alongside new data to alleviate forgetting.

Our proposed DLER belongs to the rehearsal-based category and focuses on the online setting. It exploits the complementary properties of dot-product and cosine logits for model optimization and memory management.

\subsection{Online Continual Learning}

Experience Replay (ER)~\cite{rolnick2019experience} serves as the foundation for many OCL methods, which can be broadly divided into sample management and model update strategies. Sample management strategies focus on selecting informative incoming samples for storage~\cite{aljundi2019gradient,jin2021gradient} or identifying valuable samples in the memory buffer for replay~\cite{aljundi2019online,shim2021online}. More recently, GRASP~\cite{harun2025grasp} improves rehearsal efficiency by initially prioritizing prototypical samples and progressively shifting toward more difficult ones, thereby reducing the number of model updates needed for continual learning.

Model update strategies refine the training objectives or optimization procedures for learning from incoming and buffered samples~\cite{mai2021supervised,gu2022not,wang2023cba,lin2023uer,liang2023loss,guo2022online}. SS-IL~\cite{ahn2021ss} and ER-ACE~\cite{caccia2021new} apply distinct cross-entropy objectives to these two groups of samples to alleviate prediction bias. PCR~\cite{lin2023pcr} examines their interaction through gradient propagation, whereas LODE~\cite{liang2023loss} separates the loss functions associated with new and old classes. More recent work explores alternative optimization formulations for rehearsal-based OCL. GEC~\cite{lai2025gec} formulates ER as an $\epsilon$-constraint optimization problem and adapts the update direction based on an allowable tolerance for memory loss. LOR~\cite{lai2026lor} mitigates conflicts between incoming and replay gradients by examining lookahead model states and optimizing a smooth worst objective across their trade-offs.

Building on UER~\cite{lin2023uer}, DLER extends the use of dual logits from model optimization to memory management. It introduces a supervised contrastive objective to explicitly encourage intra-class compactness and inter-class separation. Furthermore, its memory update strategy jointly uses the probability distributions produced by dot-product and cosine logits to score candidate samples and prioritize their retention.
\section{Preliminary}
\label{sec:preliminary}

OCL divides a data stream into a sequence of mini learning batches as $\mathcal{D}=\{\mathcal{B}_t\}_{t=1}^T$, where $\mathcal{B}_t=\{\mathcal{X}_t\times \mathcal{Y}_t\}$ contains the samples $\mathcal{X}_t$ and corresponding labels $\mathcal{Y}_t$. All learned classes are denoted as $\mathcal{C}_{t}$ up to step $t$. The neural network is made up of a feature extractor $\bm{h}=h_{\bm{\Phi}}(\bm{x})$ and a predictor $f(\bm{h})=\bm{W}\bm{h}^T+\bm{b}$, where $\bm{h}$ is non-negative by Relu, $\bm{W}=[\bm{w}_i]$, and $\bm{b}=[b_i]$. Since the output of predictor is dot-product logits, the categorical prediction of sample $\bm{x}$ is $p^{dot}(\bm{x})=[p_i^{dot}]$, and $p_i^{dot}$ is the probability that a sample belongs to class $i\in\mathcal{C}_{t}$:
\begin{equation}
    \label{eq:probability}
	p_i^{dot}=\frac{exp({\bm{w}_ih_{\bm{\Phi}}(\bm{x})^T+b_i})}{\sum_{j\in\mathcal{C}_{t}}exp({\bm{w}_jh_{\bm{\Phi}}(\bm{x})^T+b_j})}.
\end{equation}
During training, the model receives the current mini-batch $\mathcal{B}t$, with each sample appearing only once in the data stream. Without replaying, the training objective is
\begin{equation}
    \label{eq:oclloss}
	L_{ocl}=E_{(\bm{x},y)\sim{\mathcal{B}_t}}[l(p^{dot}(\bm{x}), \bm{y})],
\end{equation}
where $l(\cdot)$ is the cross-entropy loss function.

ER~\cite{rolnick2019experience}, as a classic rehearsal-based method, allocates a memory buffer $\mathcal{M}$ to store part of previous samples, and selects them as $\mathcal{B}_{\mathcal{M}}$ to replay with current samples. Thus, the objective function can be changed to
\begin{equation}
    \label{eq:erloss}
	L_{er}=E_{(\bm{x},y)\sim{\mathcal{B}_t\cup\mathcal{B}_{\mathcal{M}}}}[l(p^{dot}(\bm{x}), \bm{y})].
\end{equation}

LUCIR~\cite{hou2019learning} improves replay-based learning by introducing a cosine-normalized classifier. The resulting class probability distribution is denoted by $p^{\cos}(\bm{x})=[p_i^{\cos}]$, and the corresponding classification objective is 
\begin{equation}
    \label{eq:lucirloss}
    L_{lucir}=E_{(\bm{x},y)\sim{\mathcal{B}_t\cup\mathcal{B}_{\mathcal{M}}}}[l(p^{cos}(\bm{x}), \bm{y})],
\end{equation}
where 
\begin{equation}
    \label{eq:probability_cos}    p_i^{cos}=\frac{exp({cos(\bm{w}_i,h_{\bm{\Phi}}(\bm{x}))\cdot\gamma})}{\sum_{j\in\mathcal{C}_{1:t}}exp({cos(\bm{w}_j,h_{\bm{\Phi}}(\bm{x}))\cdot\gamma})}.
\end{equation}
$\gamma$ is a scale parameter and is usually set to 10.

\section{Dual-Logit Experience Replay}
\label{sec:method}

Inspired by the preceding analyses in Appendix~\ref{sec:appendix}, we propose Dual-Logit Experience Replay (DLER), which exploits cosine and dot-product logits for model optimization and memory management in OCL. Algorithm~\ref{alg:dler} summarizes this procedure, whose three components are described below.

\subsection{Online Continual Training}
\label{sec:online_training}

\subsubsection{Dual-logits training}DLER uses the cosine and dot-product predictions defined in Eqs.~\eqref{eq:probability_cos} and~\eqref{eq:probability} to construct complementary objectives. For each sample, we use an original view $\bm{x}^{(1)}=\bm{x}$ and an augmented view $\bm{x}^{(2)}=\mathcal{T}(\bm{x})$, where $\mathcal{T}$ denotes random data augmentation. For compactness, the loss averaged over two views $s\in\{\mathrm{cos},\mathrm{dot}\}$ is written as
\begin{equation}
    \ell_s(\bm{x},y)
    =
    \frac{1}{2}\sum_{v=1}^{2}
    l\bigl(p^{s}(\bm{x}^{(v)}),\bm{y}\bigr).
    \label{eq:dler_view_loss}
\end{equation}

The learning component supervises current samples through cosine logits, exploiting angular information to acquire novel knowledge. Its objective is
\begin{equation}
    L_c
    =
    \frac{1}{|\mathcal{B}_t|}
    \sum_{(\bm{x},y)\in\mathcal{B}_t}
    \ell_{\mathrm{cos}}(\bm{x},y).
    \label{eq:first_term}
\end{equation}
The replaying component incorporates additional dot-product supervision to consolidate historical knowledge by
\begin{equation}
\begin{aligned}
    L_p
    = \frac{1}{|\mathcal{B}_{\mathcal{M}}|}
    \sum_{(\bm{x},y)\in\mathcal{B}_{\mathcal{M}}}
    \Bigl[
    &\alpha\,\ell_{\mathrm{dot}}(\bm{x},y) \\
    &+(1-\alpha)\,\ell_{\mathrm{cos}}(\bm{x},y)
    \Bigr],
\end{aligned}
    \label{eq:second_term}
\end{equation}
where $\alpha\in[0,1]$ controls the relative contribution of dot-product supervision. The objective of dual-logits training is 
\begin{equation}
    L_{dual} = L_c + L_p
    \label{eq:dler_train}
\end{equation}

\subsubsection{Dual-logits enhancement}
To enhance the dual-logits mechanism, we introduce a supervised contrastive loss~\cite{mai2021supervised} over the joint current and replay samples. Specifically, $\mathcal{B}_t$ and $\mathcal{B}_{\mathcal{M}}$ are concatenated together with their labels, and both original and augmented views are processed by the shared feature extractor. Let $\mathcal{I}_t$ index all resulting views, and let $\bm{h}_i$ denote the feature of view $i$. For each anchor $i$, the comparison set is $\mathcal{A}(i)=\mathcal{I}_t\setminus\{i\}$, and the positive set is $\mathcal{P}(i)=\{j\in\mathcal{A}(i):y_j=y_i\}$. The loss is
\begin{equation}
\begin{aligned}
    L_{\mathrm{sc}}
    ={}&
    -\frac{1}{|\mathcal{I}_t|}
    \sum_{i\in\mathcal{I}_t}
    \frac{1}{|\mathcal{P}(i)|}
    \sum_{j\in\mathcal{P}(i)}
    \log
    \frac{
        \exp\bigl(\cos(\bm{h}_i,\bm{h}_j)/\tau\bigr)
    }{
        \displaystyle\sum_{a\in\mathcal{A}(i)}
        \exp\bigl(\cos(\bm{h}_i,\bm{h}_a)/\tau\bigr)
    },
\end{aligned}
\label{eq:dler_supcon}
\end{equation}
where $\tau$ is the contrastive temperature. 

\textbf{Why use the dual-logits enhancement?} Dual-logit training uses dot-product supervision during replay to help correct predictions biased toward current classes. However, the class weight norms are shared across samples, and their corrective effect depends on each sample's angular relationships with the class weights. If same-class features are widely dispersed, a correction that benefits a replay sample may not benefit other samples of that class. Encouraging angular compactness can make these shared corrections more consistent across samples.

Consider a previous class $p$ and a competing class $c$. For a feature $\bm{h}$, define their dot-product margin as
\begin{equation}
\begin{aligned}
    m_{pc}(\bm{h})
    ={}&
    \|\bm{h}\|_2
    \Bigl[
        \|\bm{w}_p\|_2\cos(\bm{w}_p,\bm{h})\\
    &\qquad-
        \|\bm{w}_c\|_2\cos(\bm{w}_c,\bm{h})
    \Bigr]
    +b_p-b_c.
\end{aligned}
\label{eq:sc_dot_margin}
\end{equation}
A positive margin means that dot-product prediction favors class $p$ over class $c$, even when their cosine similarities have the opposite ordering. To examine how this margin transfers between samples within the same class, let $\bm{h}_r$ be the feature of a replay sample and $\bm{h}_u$ the feature of another sample from class $p$ that is not included in the current replay mini-batch. Both features are evaluated under the same model. For nonzero features, define the normalized margin
\begin{equation}
    \widetilde{m}_{pc}(\bm{h})
    =
    \frac{m_{pc}(\bm{h})}{\|\bm{h}\|_2}.
\label{eq:sc_normalized_margin}
\end{equation}
Their margin difference can be expressed as
\begin{equation}
\begin{aligned}
    &\widetilde{m}_{pc}(\bm{h}_u)
    -\widetilde{m}_{pc}(\bm{h}_r)\\
    &=
    (\bm{w}_p-\bm{w}_c)
    \left(
        \frac{\bm{h}_u}{\|\bm{h}_u\|_2}
        -
        \frac{\bm{h}_r}{\|\bm{h}_r\|_2}
    \right)^T\\
    &\quad+
    (b_p-b_c)
    \left(
        \frac{1}{\|\bm{h}_u\|_2}
        -
        \frac{1}{\|\bm{h}_r\|_2}
    \right).
\end{aligned}
\label{eq:sc_margin_difference}
\end{equation}
Applying the Cauchy--Schwarz inequality~\cite{steele2004cauchy} gives
\begin{equation}
\begin{aligned}
    \widetilde{m}_{pc}(\bm{h}_u)
    \geq{}&
    \widetilde{m}_{pc}(\bm{h}_r)\\
    &-
    \|\bm{w}_p-\bm{w}_c\|_2
    \sqrt{2\bigl[1-\cos(\bm{h}_u,\bm{h}_r)\bigr]}\\
    &-
    |b_p-b_c|
    \left|
        \frac{1}{\|\bm{h}_u\|_2}
        -
        \frac{1}{\|\bm{h}_r\|_2}
    \right|.
\end{aligned}
\label{eq:sc_margin_bound}
\end{equation}

Equation~\eqref{eq:sc_margin_bound} shows that, with the remaining quantities fixed, greater same-class cosine similarity reduces the angular discrepancy term. If the replay sample has a positive margin exceeding the two discrepancy terms, the other sample also has a positive margin against class $c$. The bias-dependent term vanishes when $b_p=b_c$ and becomes small when the feature norms are similar, while correct classification requires a positive margin against every competing class. These observations suggest that within-class angular compactness can support dual-logit correction by allowing previous samples with similar directions to benefit more consistently from the same norm factor. The supervised contrastive objective encourages this geometry using current and replay samples, potentially extending corrections learned from limited replay data to more samples of the same class.

\begin{algorithm}[t]
\caption{Dual-Logit Experience Replay}
\label{alg:dler}
\renewcommand{\algorithmicrequire}{\textbf{Input:}}
\renewcommand{\algorithmicensure}{\textbf{Output:}}
\begin{algorithmic}[1]
\REQUIRE Stream $\{\mathcal{B}_t\}_{t=1}^{T}$, memory capacity $M$, replay size $b_m$, learning rate $\eta$, and parameters $\alpha,\lambda_{\mathrm{sc}},\gamma,\tau$
\ENSURE Network parameters $\bm{\theta}$ and memory buffer $\mathcal{M}$
\STATE Initialize $\bm{\theta}$, $\mathcal{M}\leftarrow\varnothing$, $\mathcal{C}\leftarrow\varnothing$, and class counts $N_c\leftarrow0$
\FOR{$t=1,\ldots,T$}
    \STATE $\mathcal{C}\leftarrow\mathcal{C}\cup\{y:(\bm{x},y)\in\mathcal{B}_t\}$
    \STATE Uniformly sample up to $b_m$ examples without replacement from $\mathcal{M}$ to obtain $\mathcal{B}_{\mathcal{M}}$
    \STATE Construct original and augmented views of current and replay samples
    \STATE Compute $L_c$ using Eq.~\eqref{eq:first_term}
    \IF{$\mathcal{B}_{\mathcal{M}}\neq\varnothing$}
        \STATE Compute $L_p$ using Eq.~\eqref{eq:second_term}
        \STATE Compute joint $L_{\mathrm{sc}}$ using Eq.~\eqref{eq:dler_supcon}
        \STATE $L_{dler}\leftarrow L_c+L_p+\lambda_{\mathrm{sc}}L_{\mathrm{sc}}$
    \ELSE
        \STATE $L_{dler}\leftarrow L_c$
    \ENDIF
    \STATE $\bm{\theta}\leftarrow\bm{\theta}-\eta\nabla_{\bm{\theta}}L$
    \STATE Update $\mathcal{M}$ and $N$ using Algorithm~\ref{alg:dler_memory}
\ENDFOR
\STATE Predict test labels using Eq.~\eqref{eq:dler_test}
\end{algorithmic}
\end{algorithm}

\subsubsection{The overall objective} Based on the dual-logits training and its enhancement, the overall objective of online continual training is thus denoted as
\begin{equation}
    L_{dler} = L_{c} + L_{p}+\lambda_{\mathrm{sc}}L_{\mathrm{sc}},
    \label{eq:dler_train}
\end{equation}
where $\lambda_{\mathrm{sc}}$ controls the auxiliary loss. When memory is empty, only $L_c$ is used. The model is updated by
\begin{equation}
    \bm{\theta}
    \leftarrow
    \bm{\theta}-\eta\nabla_{\bm{\theta}}L,
    \label{eq:dler_update}
\end{equation}
where $\bm{\theta}=\{\bm{\Phi},\bm{W},\bm{b}\}$ and $\eta>0$ is the learning rate.

\subsection{Online Continual Testing}
\label{sec:online_testing}

During testing, DLER uses the dot-product prediction defined in Eq.~\eqref{eq:probability} to incorporate both angular and norm factors. For each test sample $\bm{x}$, the predicted label is
\begin{equation}
\hat{y}
=
\underset{c\in\mathcal{C}_t}{\arg\max}\quad p_c^{dot}(\bm{x}),
\label{eq:dler_test}
\end{equation}
which is restricted to the classes observed up to step $t$.

\subsection{Memory buffer Management}
\label{sec:memory_management}

\subsubsection{Random memory retrieval} The memory buffer stores at most $M$ original samples, regardless of the length of the data stream. Before each model update, up to $b_m$ historical samples are retrieved uniformly without replacement to form $\mathcal{B}_{\mathcal{M}}$.

\subsubsection{Dual-logits memory update} After optimization, DLER updates the memory buffer using dual-logits. The update consists of balanced memory allocation, dual-logits scoring, and score-guided replacement, as summarized in Algorithm~\ref{alg:dler_memory}.

\begin{algorithm}[t]
\caption{Dual-Logit Memory Update}
\label{alg:dler_memory}
\renewcommand{\algorithmicrequire}{\textbf{Input:}}
\renewcommand{\algorithmicensure}{\textbf{Output:}}
\begin{algorithmic}[1]
\REQUIRE Updated model $\bm{\theta}$, memory $\mathcal{M}$,
mini-batch $\mathcal{B}_t$, observed classes $\mathcal{C}_t$,
capacity $M$
\ENSURE Updated memory $\mathcal{M}$

\STATE Compute $\{q_c^{(t)}\}_{c\in\mathcal{C}_t}$
using Eq.~\eqref{eq:dler_quota}
\STATE Score samples in $\mathcal{M}$ and $\mathcal{B}_t$
using Eq.~\eqref{eq:dler_score}

\FOR{each $c\in\mathcal{C}_t$}
    \STATE Retain the top
    $\min\{q_c^{(t)},|\mathcal{M}_c|\}$
    samples in $\mathcal{M}_c$ by score
\ENDFOR

\FOR{each $(\bm{x},y)\in\mathcal{B}_t$ sequentially}
    \IF{$q_y^{(t)}=0$}
        \STATE \textbf{continue}
    \ENDIF
    \IF{$|\mathcal{M}_y|<q_y^{(t)}$}
        \STATE $\mathcal{M}_y \leftarrow
        \mathcal{M}_y\cup\{(\bm{x},y)\}$
    \ELSE
        \STATE Select $(\bm{x}^{-},y)$ using
        Eq.~\eqref{eq:dler_victim}
        \IF{$s(\bm{x},y)>s(\bm{x}^{-},y)$}
            \STATE Replace $(\bm{x}^{-},y)$ with $(\bm{x},y)$
            using Eq.~\eqref{eq:dler_replacement}
        \ENDIF
    \ENDIF
\ENDFOR
\end{algorithmic}
\end{algorithm}

\textbf{Balanced Memory Allocation.} Let $K_t=|\mathcal{C}_t|$ denote the number of observed classes and $r_t=M-K_t\lfloor M/K_t\rfloor$ the capacity after equal allocation. The quota for class $c$ is
\begin{equation}
    q_c^{(t)}
    =
    \left\lfloor\frac{M}{K_t}\right\rfloor
    +
    \mathbb{I}\!\left[
    \operatorname{rank}_t(c)\leq r_t
    \right],
    \label{eq:dler_quota}
\end{equation}
where $\operatorname{rank}_t(c)$ denotes the position of class $c$ among the observed classes sorted by label index. The quotas sum to $M$. When newly observed classes reduce an existing class quota, we retain its highest-scoring samples up to the updated quota.

\textbf{Dual-Logit Scoring.} The scoring criterion follows from the complementary roles of cosine and dot-product logits in replay. Cosine predictions rely on angular information and may remain biased toward current classes. Dot-product predictions additionally incorporate norm information, which can increase the probability assigned to the correct previous class. We therefore prioritize samples that exhibit this positive probability advantage, aiming to preserve examples that support the correction mechanism of dual-logit training. For a labeled sample $(\bm{x},y)$, the probability advantage is
\begin{equation}
    a(\bm{x},y)
    =
    \left[
    p_y^{\mathrm{dot}}(\bm{x})
    -
    p_y^{\mathrm{cos}}(\bm{x})
    \right]_+,
    \label{eq:dler_advantage}
\end{equation} 
where $[v]_+=\max(v,0)$ and both probabilities correspond to the true class $y$. We directly use this advantage as the score 
\begin{equation}
    s(\bm{x},y)=a(\bm{x},y).
    \label{eq:dler_score}
\end{equation}
Since both probabilities lie in $[0,1]$, the score belongs
to $[0,1]$.

\textbf{Score-Guided Replacement.} After optimization, we rescore the original samples in
$\mathcal{M}$ and $\mathcal{B}_t$ using the updated model.
For a sample $(\bm{x},y)$, no insertion is performed if $q_y^{(t)}=0$. If $|\mathcal{M}_y|<q_y^{(t)}$, the sample is inserted directly. Otherwise, we select a sample stored in the memory buffer with the lowest score from the same class by
\begin{equation}
    (\bm{x}^{-},y)
    \in
    \underset{(\bm{x}',y)\in\mathcal{M}_y}
    {\arg\min}\;s(\bm{x}',y),
    \label{eq:dler_victim}
\end{equation}
with ties resolved uniformly at random. Replacement occurs only when the incoming sample has a higher score, as
$s(\bm{x},y)>s(\bm{x}^{-},y)$, and
\begin{equation}
    \mathcal{M}_y \leftarrow
    \bigl(\mathcal{M}_y\setminus\{(\bm{x}^{-},y)\}\bigr)
    \cup\{(\bm{x},y)\}.
    \label{eq:dler_replacement}
\end{equation}
Otherwise, $\mathcal{M}_y$ remains unchanged. This procedure uses the same predictive advantage to guide both quota reduction and sample replacement. Rescoring requires additional forward passes during training, which can be processed in mini-batches, but introduces no additional inference modules or test-time computation. 
As summarized in Algorithm~\ref{alg:dler}, memory retrieval precedes model optimization, and memory updates follow it. 

\textbf{Why use the dual-logits memory update?} Dual-logit training incorporates additional dot-product supervision during replay, but reservoir sampling does not consider which samples benefit from this prediction mechanism. When only a few samples can be retained per class, we prioritize those for which dot-product prediction assigns a higher probability to the correct class than cosine prediction. Retaining such samples provides opportunities to exploit norm information during replay, potentially reducing the reliance on angular adjustments to preserve historical predictions. This may help limit interference with the angular learning of current samples through cosine supervision. Scoring with the latest model identifies this predictive advantage under competition among the currently observed classes, aligning limited replay memory with the dual-logit training mechanism.

\begin{table*}[!t]
\centering
\caption{Final Average Accuracy (\%). Best and second-best available results among continual learning methods, excluding UER, are boldfaced and underlined, respectively. }
\label{tab:final-accuracy}
\vspace{-1ex}
\begingroup
\footnotesize
\setlength{\tabcolsep}{3pt}
\renewcommand{\arraystretch}{1.15}

\begin{adjustbox}{max width=\textwidth}
\begin{tabular}{@{}l
    *{4}{c}@{\hspace{8pt}}
    *{4}{c}@{\hspace{8pt}}
    *{4}{c}@{}}
\toprule
\textbf{Method}
& \multicolumn{4}{c}{\textbf{Split CIFAR100}}
& \multicolumn{4}{c}{\textbf{Split MiniImageNet}}
& \multicolumn{4}{c}{\textbf{Split TinyImageNet}} \\
\cmidrule(lr){2-5}
\cmidrule(lr){6-9}
\cmidrule(l){10-13}
Buffer size
& 100 & 200 & 500 & 1000
& 200 & 500 & 1000 & 2000
& 400 & 1000 & 2000 & 4000 \\
\midrule

IID
& \multicolumn{4}{c}{\tmmacc{17.3}{0.8}}
& \multicolumn{4}{c}{\tmmacc{18.2}{1.1}}
& \multicolumn{4}{c}{\tmmacc{17.3}{1.7}} \\

IID++~\cite{caccia2021new}
& \multicolumn{4}{c}{\tmmacc{23.5}{0.8}}
& \multicolumn{4}{c}{\tmmacc{20.7}{1.0}}
& \multicolumn{4}{c}{\tmmacc{19.1}{1.3}} \\

Fine-tune
& \multicolumn{4}{c}{\tmmacc{5.9}{0.2}}
& \multicolumn{4}{c}{\tmmacc{4.3}{0.2}}
& \multicolumn{4}{c}{\tmmacc{4.3}{0.2}} \\
\midrule

ER~\cite{rolnick2019experience}
& \tmmacc{9.0}{0.4} & \tmmacc{11.0}{0.8}
& \tmmacc{14.5}{0.8} & \tmmacc{17.6}{0.9}
& \tmmacc{8.8}{0.4} & \tmmacc{11.2}{0.6}
& \tmmacc{13.4}{0.9} & \tmmacc{16.5}{0.9}
& \tmmacc{5.2}{0.6} & \tmmacc{7.6}{0.5}
& \tmmacc{10.1}{0.9} & \tmmacc{12.7}{1.1} \\

GSS~\cite{aljundi2019gradient}
& \tmmacc{8.5}{0.8} & \tmmacc{10.4}{0.6}
& \tmmacc{14.6}{1.3} & \tmmacc{16.9}{1.4}
& \tmmacc{8.5}{0.7} & \tmmacc{10.3}{1.5}
& \tmmacc{13.9}{1.0} & \tmmacc{14.6}{1.1}
& \tmmacc{5.1}{0.6} & \tmmacc{7.0}{0.5}
& \tmmacc{10.1}{0.5} & \tmmacc{11.7}{0.6} \\

MIR~\cite{aljundi2019online}
& \tmmacc{8.9}{0.5} & \tmmacc{10.6}{0.7}
& \tmmacc{14.8}{0.7} & \tmmacc{18.1}{0.7}
& \tmmacc{8.8}{0.6} & \tmmacc{11.9}{0.6}
& \tmmacc{14.8}{1.1} & \tmmacc{17.2}{0.8}
& \tmmacc{4.9}{0.4} & \tmmacc{7.3}{0.7}
& \tmmacc{9.2}{1.3} & \tmmacc{11.3}{1.2} \\

ASER~\cite{shim2021online}
& \tmmacc{7.5}{0.8} & \tmmacc{8.8}{0.6}
& \tmmacc{13.0}{0.9} & \tmmacc{16.1}{1.1}
& \tmmacc{7.0}{0.5} & \tmmacc{10.5}{1.1}
& \tmmacc{13.8}{0.9} & \tmmacc{16.1}{1.1}
& \tmmacc{4.9}{0.3} & \tmmacc{6.9}{0.7}
& \tmmacc{8.8}{0.9} & \tmmacc{12.3}{0.8} \\

GMED~\cite{jin2021gradient}
& \tmmacc{8.2}{1.9} & \tmmacc{10.8}{2.6} & \tmmacc{15.0}{0.9} & \tmmacc{18.8}{0.7}
& \tmmacc{9.3}{0.7} & \tmmacc{11.9}{1.2}
& \tmmacc{15.3}{1.3} & \tmmacc{18.0}{0.8}
& \tmmacc{5.8}{0.4} & \tmmacc{8.2}{0.8}
& \tmmacc{9.9}{2.5} & \tmmacc{9.9}{4.6} \\

A-GEM~\cite{chaudhry2018efficient}
& \tmmacc{5.4}{0.4} & \tmmacc{5.3}{0.5}
& \tmmacc{5.4}{0.6} & \tmmacc{5.6}{0.5}
& \tmmacc{4.9}{0.7} & \tmmacc{5.0}{1.0}
& \tmmacc{4.7}{1.1} & \tmmacc{5.0}{2.3}
& \tmmacc{2.5}{0.2} & \tmmacc{2.6}{0.2}
& \tmmacc{2.5}{0.2} & \tmmacc{2.6}{0.1} \\

LUCIR~\cite{hou2019learning}
& \tmmacc{11.1}{0.4} & \tmmacc{13.4}{0.9}
& \tmmacc{14.5}{0.4} & \tmmacc{17.4}{0.8}
& \tmmacc{10.0}{0.8} & \tmmacc{12.4}{0.6}
& \tmmacc{14.7}{0.9} & \tmmacc{16.9}{1.1}
& \tmmacc{6.1}{0.4} & \tmmacc{6.7}{0.4}
& \tmmacc{8.3}{0.6} & \tmmacc{9.8}{0.6} \\

ER-WA~\cite{zhao2020maintaining}
& \tmmacc{10.9}{0.5} & \tmmacc{14.6}{0.9}
& \tmmacc{18.3}{0.7} & \tmmacc{21.7}{1.2}
& \tmmacc{11.0}{0.8} & \tmmacc{15.1}{0.7}
& \tmmacc{17.1}{0.9} & \tmmacc{18.9}{1.4}
& \tmmacc{7.9}{0.6} & \tmmacc{11.5}{0.7}
& \tmmacc{14.3}{0.9} & \tmmacc{16.4}{1.0} \\

DER++~\cite{buzzega2020dark}
& \tmmacc{10.5}{0.9} & \tmmacc{11.6}{0.7}
& \tmmacc{15.5}{1.0} & \tmmacc{17.2}{1.1}
& \tmmacc{9.0}{0.7} & \tmmacc{11.9}{1.0}
& \tmmacc{14.8}{0.7} & \tmmacc{16.1}{1.3}
& \tmmacc{5.1}{0.5} & \tmmacc{7.0}{0.9}
& \tmmacc{9.3}{0.7} & \tmmacc{10.8}{1.0} \\

SS-IL~\cite{ahn2021ss}
& \tmmsecond{13.0}{0.6} & \tmmsecond{16.1}{1.0}
& \tmmacc{19.5}{0.6} & \tmmacc{21.9}{1.1}
& \tmmsecond{14.9}{0.8} & \tmmacc{18.0}{0.7}
& \tmmacc{19.7}{0.9} & \tmmacc{21.7}{1.0}
& \tmmsecond{11.2}{0.8} & \tmmsecond{14.1}{0.5}
& \tmmsecond{17.1}{0.8} & \tmmsecond{19.6}{0.6} \\

SCR~\cite{mai2021supervised}
& \tmmacc{7.3}{0.6} & \tmmacc{8.6}{0.4}
& \tmmacc{13.3}{0.6} & \tmmacc{16.2}{1.3}
& \tmmacc{7.7}{0.6} & \tmmacc{12.1}{0.7}
& \tmmacc{14.7}{1.9} & \tmmacc{16.8}{0.6}
& \tmmacc{5.2}{0.2} & \tmmacc{7.3}{0.2}
& \tmmacc{8.7}{0.4} & \tmmacc{10.2}{0.3} \\

ER-ACE~\cite{caccia2021new}
& \tmmacc{12.1}{0.7} & \tmmacc{15.7}{0.7}
& \tmmsecond{19.7}{0.8} & \tmmsecond{23.1}{0.8}
& \tmmacc{14.6}{0.8} & \tmmsecond{18.1}{0.5}
& \tmmsecond{20.3}{1.3} & \tmmsecond{24.8}{1.1}
& \tmmacc{9.3}{0.4} & \tmmacc{12.2}{0.6}
& \tmmacc{14.6}{0.8} & \tmmacc{17.9}{0.7} \\

ER-DVC~\cite{gu2022not}
& \tmmacc{10.6}{0.5} & \tmmacc{12.9}{0.7}
& \tmmacc{16.8}{0.8} & \tmmacc{19.7}{0.7}
& \tmmacc{10.1}{0.6} & \tmmacc{13.9}{0.6}
& \tmmacc{15.4}{0.7} & \tmmacc{17.2}{0.8}
& \tmmacc{6.6}{0.6} & \tmmacc{9.4}{0.6}
& \tmmacc{11.4}{0.8} & \tmmacc{14.1}{1.0} \\

OBC~\cite{chrysakisonline}
& \tmmacc{9.5}{0.6} & \tmmacc{13.3}{0.9}
& \tmmacc{18.7}{0.9} & \tmmacc{22.1}{0.6}
& \tmmacc{9.2}{0.3} & \tmmacc{14.4}{0.9}
& \tmmacc{16.4}{1.4} & \tmmacc{19.5}{1.5}
& \tmmacc{7.3}{0.4} & \tmmacc{11.0}{0.7}
& \tmmacc{13.2}{0.6} & \tmmacc{16.2}{1.0} \\


LOR~\cite{lai2026lor}
& \tmmacc{9.5}{0.5} & \tmmacc{11.6}{0.7}
& \tmmacc{15.8}{0.9} & \tmmacc{19.2}{0.8}
& \tmmacc{9.5}{1.0} & \tmmacc{11.2}{0.6}
& \tmmacc{15.2}{2.2} & \tmmacc{19.1}{1.8}
& \tmmacc{5.5}{0.4} & \tmmacc{8.6}{0.3}
& \tmmacc{11.9}{0.9} & \tmmacc{13.9}{1.0} \\
\midrule

UER~\cite{lin2023uer}
& \tmmacc{12.5}{0.6} & \tmmacc{15.4}{0.7}
& \tmmacc{20.9}{0.8} & \tmmacc{24.6}{0.8}
& \tmmacc{12.8}{0.9} & \tmmacc{18.4}{0.8}
& \tmmacc{21.9}{1.3} & \tmmacc{25.1}{1.1}
& \tmmacc{9.7}{0.4} & \tmmacc{13.4}{0.5}
& \tmmacc{16.2}{0.7} & \tmmacc{18.9}{0.8} \\

\textbf{DLER}
& \tmmbest{14.4}{0.5} & \tmmbest{18.7}{0.5}
& \tmmbest{23.6}{0.8} & \tmmbest{25.1}{1.2}
& \tmmbest{15.0}{0.9} & \tmmbest{20.2}{1.3}
& \tmmbest{23.1}{1.0} & \tmmbest{25.6}{0.7}
& \tmmbest{11.9}{0.3} & \tmmbest{15.8}{0.5}
& \tmmbest{18.5}{0.5} & \tmmbest{20.3}{0.6} \\

\bottomrule
\end{tabular}
\end{adjustbox}
\endgroup
\vspace{-4ex}
\end{table*}

\section{Experiments}
\label{sec:experiments}

\subsection{Experiment Setup}
\subsubsection{Datasets}

We conduct extensive experiments on three publicly available datasets. \textbf{Split CIFAR100}~\cite{krizhevsky2009learning} and \textbf{Split MiniImageNet}~\cite{vinyals2016matching} are made up of 10 learning stages, and each stage has 10 classes.\textbf{Split TinyImageNet}~\cite{le2015tiny} consists of 20 learning stages, and each stage contains 10 classes.

\subsubsection{Evaluation Metrics}
We first define $a_{i,j}(j<=i)$ as the accuracy evaluated on the held-out test samples of the $j$th task after the network has learned the first $i$ tasks. Similar to~\cite {shim2021online}, we can further acquire the average accuracy rate
\begin{equation}
\label{eq:first}
 	\begin{split}
	 	A_i = \frac{1}{i}\sum_{j=1}^{i}a_{i,j},
 	\end{split}
\end{equation}
at the $i$th task. The final accuracy rate for $T$ tasks is $A_T$.

\subsubsection{Implementation Details}

The basic setting of the backbone model is the same as the existing work~\cite{lai2026lor,lai2025gec}. In detail, we take the Reduced ResNet18 (the number of filters is 20) as the feature extractor on all datasets. All of the networks are randomly initialized rather than pre-trained. During training, the network is trained with the SGD optimizer, and the learning rate is set to 0.1. Meanwhile, we select $\alpha$ and $\lambda_{sc}$ on a validation set that is obtained by sampling 10\% of the training set. As for testing, we select 256 as the batch size. For all datasets, the class indexes are shuffled before division. The model receives 10 current samples from the data stream and 10 previous samples from the memory buffer at a time. We also combine multiple data augmentation techniques to generate augmented views of the training samples.

\begin{figure*}[t]
\centering
\subfigure[Split CIFAR100]{
    \begin{minipage}[t]{0.31\linewidth}
        \centering
        \includegraphics[scale=0.26]{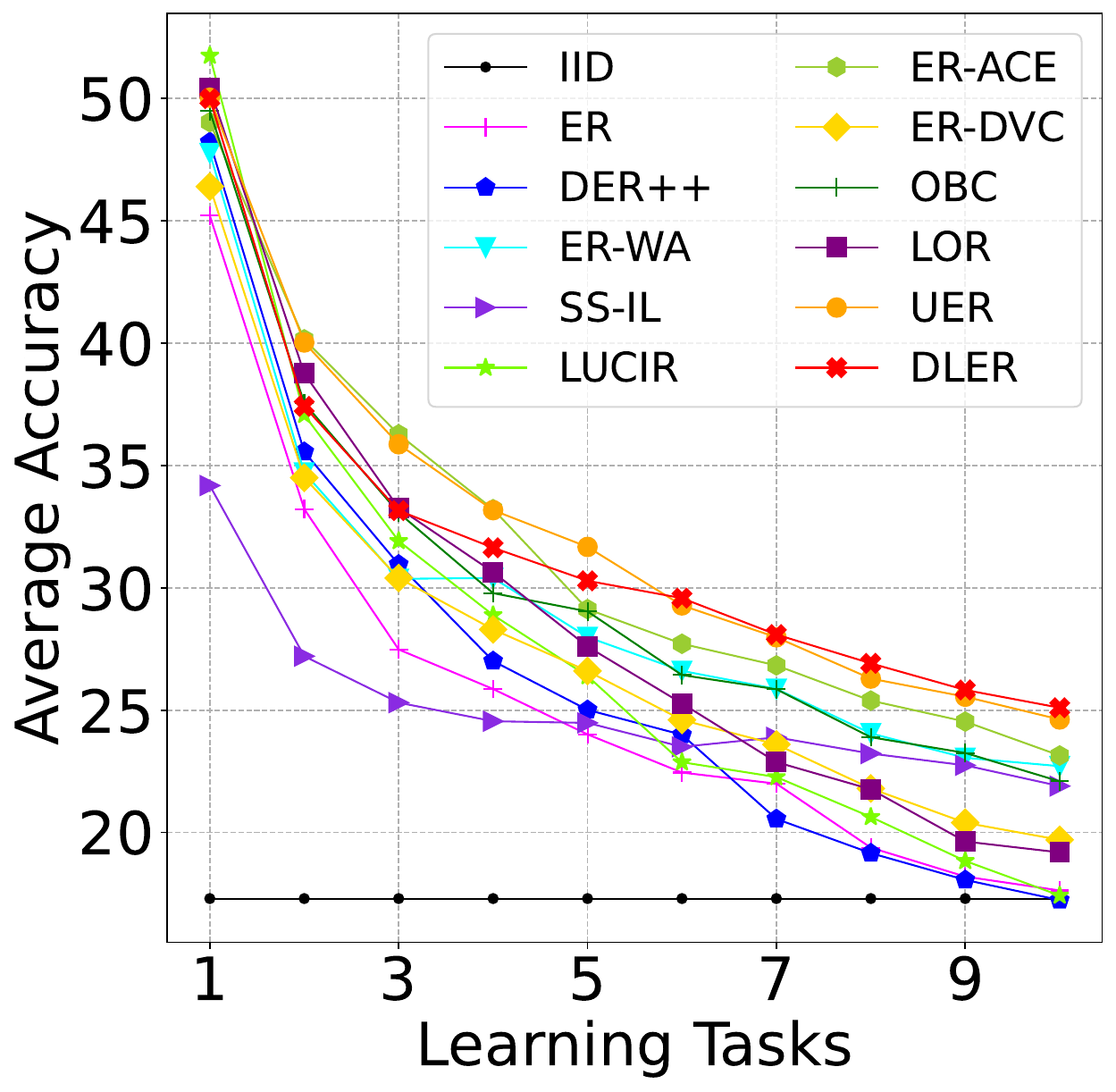}
    \end{minipage}
}
\subfigure[Split MiniImageNet]{
    \begin{minipage}[t]{0.31\linewidth}
        \centering
        \includegraphics[scale=0.26]{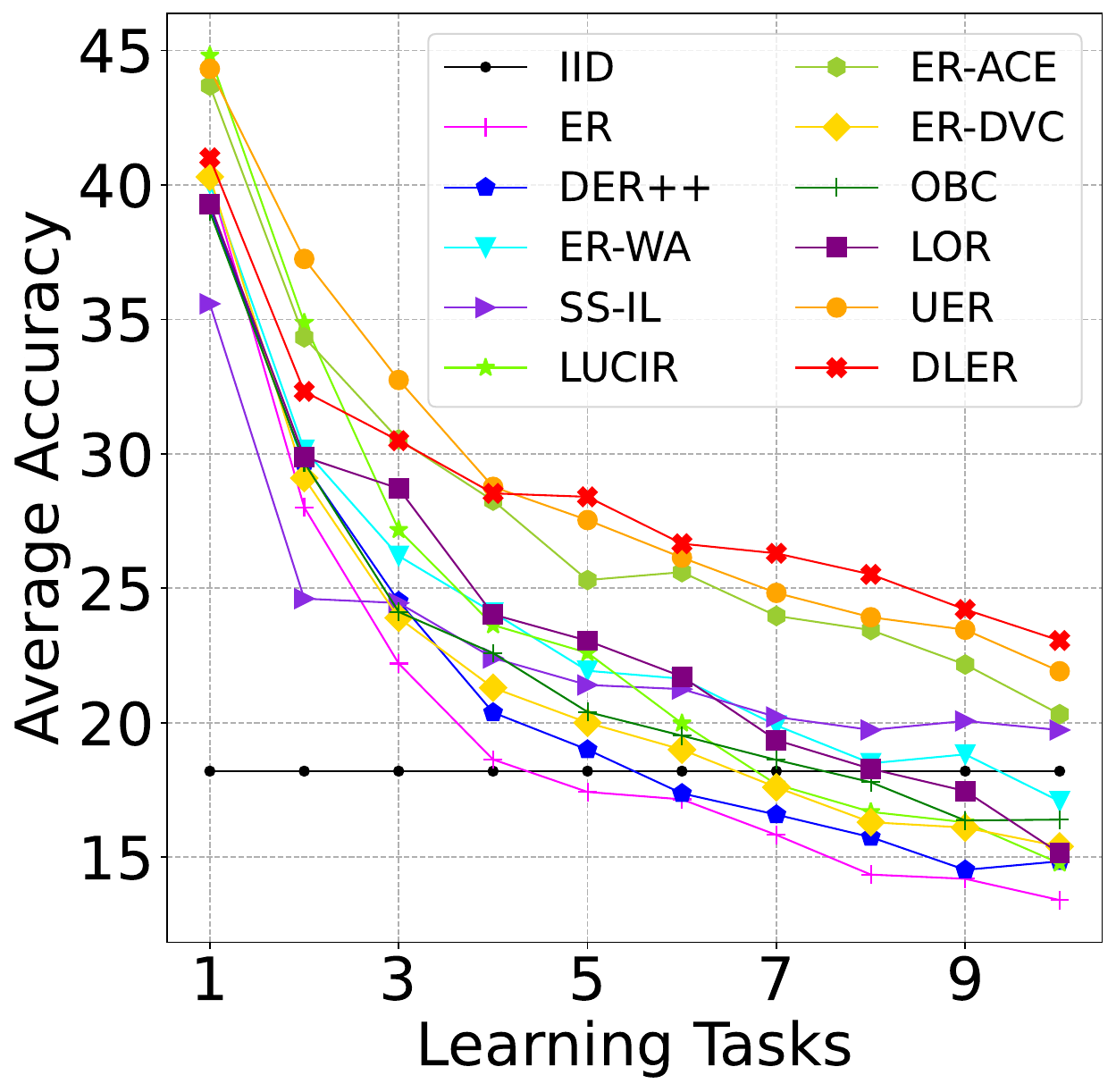}
    \end{minipage}
}
\subfigure[Split TinyImageNet]{
    \begin{minipage}[t]{0.31\linewidth}
        \centering
        \includegraphics[scale=0.26]{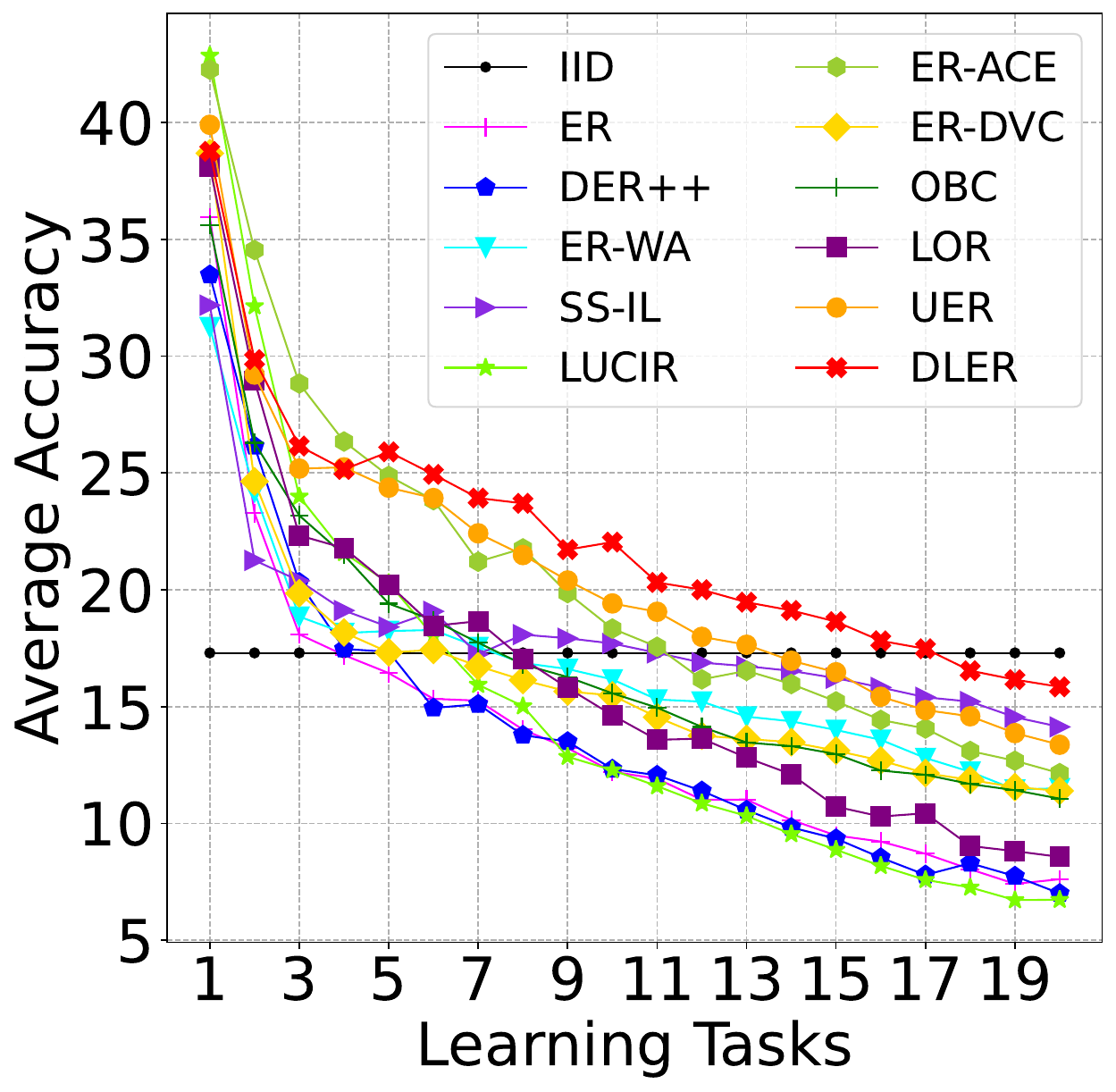}
    \end{minipage}
}
\centering
\vspace{-2ex}
\caption{Average accuracy rate on observed learning stages on three datasets when the memory buffer size is 1000. }
\label{learningstage1000}
\vspace{-2ex}
\end{figure*}

\subsection{Overall Performance}

We evaluate DLER against state-of-the-art baselines. Table~\ref{tab:final-accuracy} reports the final average accuracy as the mean $\pm$ the half-width of the 95\% confidence interval over 10 runs, using distinct, randomly generated class orders for each dataset. 

DLER achieves the highest mean accuracy across all 12 experimental settings. On Split CIFAR100, DLER outperforms the strongest competing baseline in each setting by 1.4\%, 2.6\%, 3.9\%, and 2.0\% with buffer sizes of 100, 200, 500, and 1000, respectively. On Split MiniImageNet, the corresponding improvements are 0.1\%, 2.1\%, 2.8\%, and 0.8\% with buffer sizes of 200, 500, 1000, and 2000. On Split TinyImageNet, DLER consistently outperforms SS-IL by 0.7\%, 1.7\%, 1.4\%, and 0.7\% with buffer sizes of 400, 1000, 2000, and 4000, respectively. These results demonstrate the effectiveness of DLER across different datasets and memory capacities, including settings with limited memory.

Compared with UER, DLER improves mean accuracy in every evaluated setting, with gains ranging from 0.5\% to 3.3\%. Notably, the gains on Split TinyImageNet remain above one percentage point across all four memory capacities, demonstrating the benefits of the extensions under a longer learning sequence with more classes. These results support the combined effectiveness of enhancing dual-logit training through improved feature compactness and memory management.

Figure~\ref{learningstage1000} reports accuracy after each learning stage. Although DLER does not achieve the highest accuracy initially, it leads the compared methods from stage 6 onward on Split CIFAR100 and from stage 5 onward on Split MiniImageNet and Split TinyImageNet. These gains suggest an improved balance between acquiring novel knowledge and retaining historical knowledge as more classes share the fixed memory.

\begin{table*}[t]
\centering
\caption{Ablation study on Split TinyImageNet with different memory buffer sizes. }
\label{tab:ablation_tinyimagenet}
\vspace{-1ex}
\begingroup
\small
\setlength{\tabcolsep}{3pt}
\renewcommand{\arraystretch}{1.15}

\begin{tabular*}{\textwidth}{@{\extracolsep{\fill}}l*{9}{c}@{}}
\toprule
& \multicolumn{5}{c}{Components}
& \multicolumn{4}{c}{Buffer size} \\
\cmidrule(lr){2-6}
\cmidrule(l){7-10}
Method
& \shortstack{Dual-logits}
& \shortstack{$\alpha$}
& SC & BM & Advantage
& 400 & 1000 & 2000 & 4000 \\
\midrule

ER
& & & & &
& \tmmacc{5.2}{0.6} & \tmmacc{7.6}{0.5}
& \tmmacc{10.1}{0.9} & \tmmacc{12.7}{1.1} \\

Dual-logit training (Fixed)
& $\checkmark$ & & & &
& \tmmacc{8.0}{0.6} & \tmmacc{10.9}{0.7}
& \tmmacc{14.2}{0.8} & \tmmacc{16.1}{0.8} \\

Dual-logit training
& $\checkmark$ & $\checkmark$ & & &
& \tmmacc{9.7}{0.4} & \tmmacc{13.4}{0.5}
& \tmmacc{16.2}{0.7} & \tmmacc{18.9}{0.8} \\

Dual-logit training + SC
& $\checkmark$ & $\checkmark$ & $\checkmark$ & &
& \tmmacc{10.0}{0.5} & \tmmacc{13.9}{0.3}
& \tmmacc{17.2}{0.5} & \tmmacc{19.6}{0.7} \\

Dual-logit training + SC + CB
& $\checkmark$ & $\checkmark$ & $\checkmark$ & $\checkmark$ &
& \tmmacc{10.2}{0.5} & \tmmacc{14.5}{0.3}
& \tmmacc{17.4}{0.8} & \tmmacc{19.7}{0.5} \\

\midrule
\textbf{DLER}
& $\checkmark$ & $\checkmark$ & $\checkmark$
& $\checkmark$ & $\checkmark$
& \tmmbest{11.9}{0.3} & \tmmbest{15.8}{0.5}
& \tmmbest{18.5}{0.5} & \tmmbest{20.3}{0.6} \\

\bottomrule
\end{tabular*}
\endgroup
\vspace{-3ex}
\end{table*}

\begin{figure*}[t]
\centering
\subfigure[Previous Classes]{
    \begin{minipage}[t]{0.31\linewidth}
        \centering
        \includegraphics[scale=0.26]{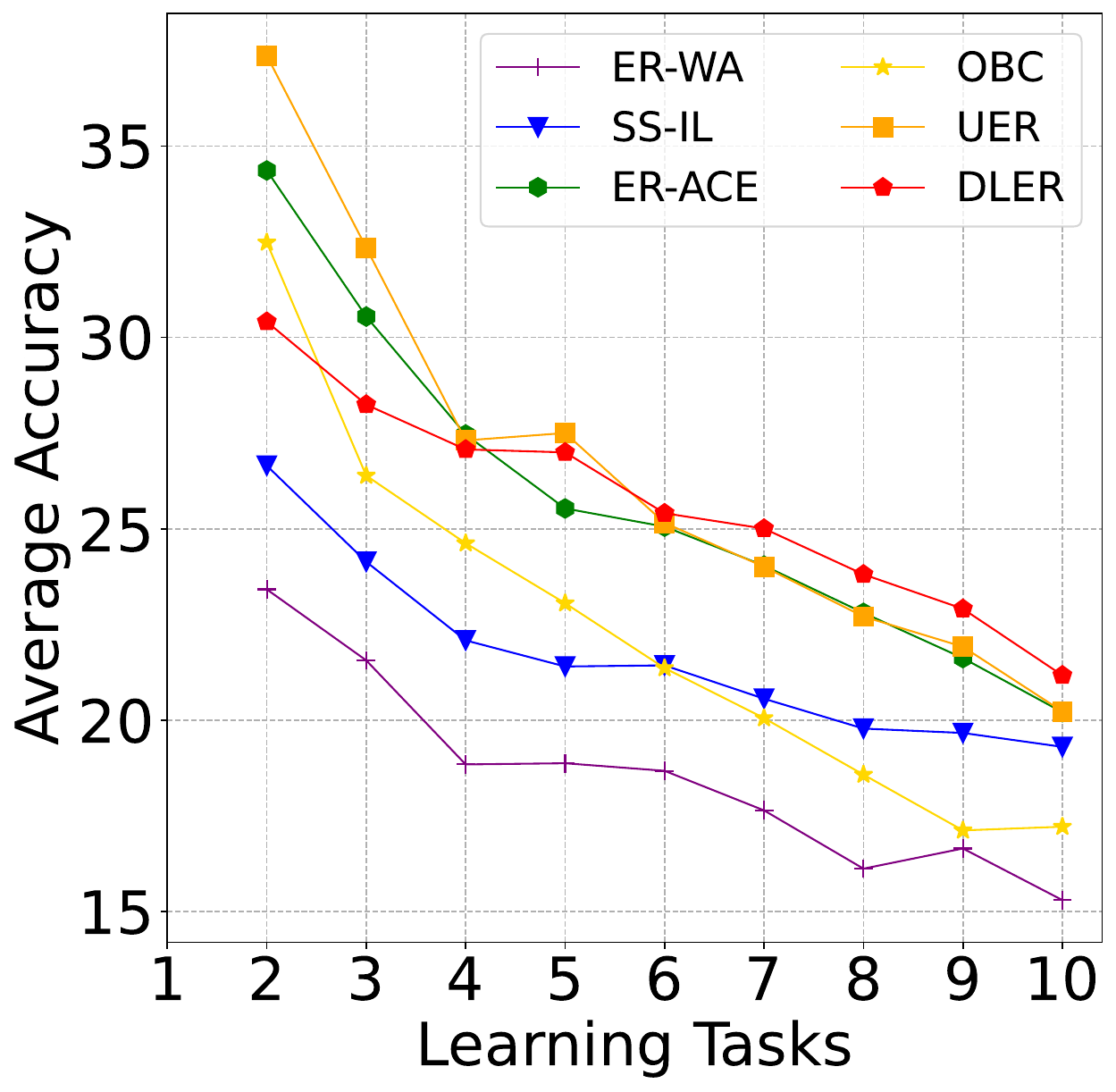}
    \end{minipage}
}
\subfigure[Current Classes]{
    \begin{minipage}[t]{0.31\linewidth}
        \centering
        \includegraphics[scale=0.26]{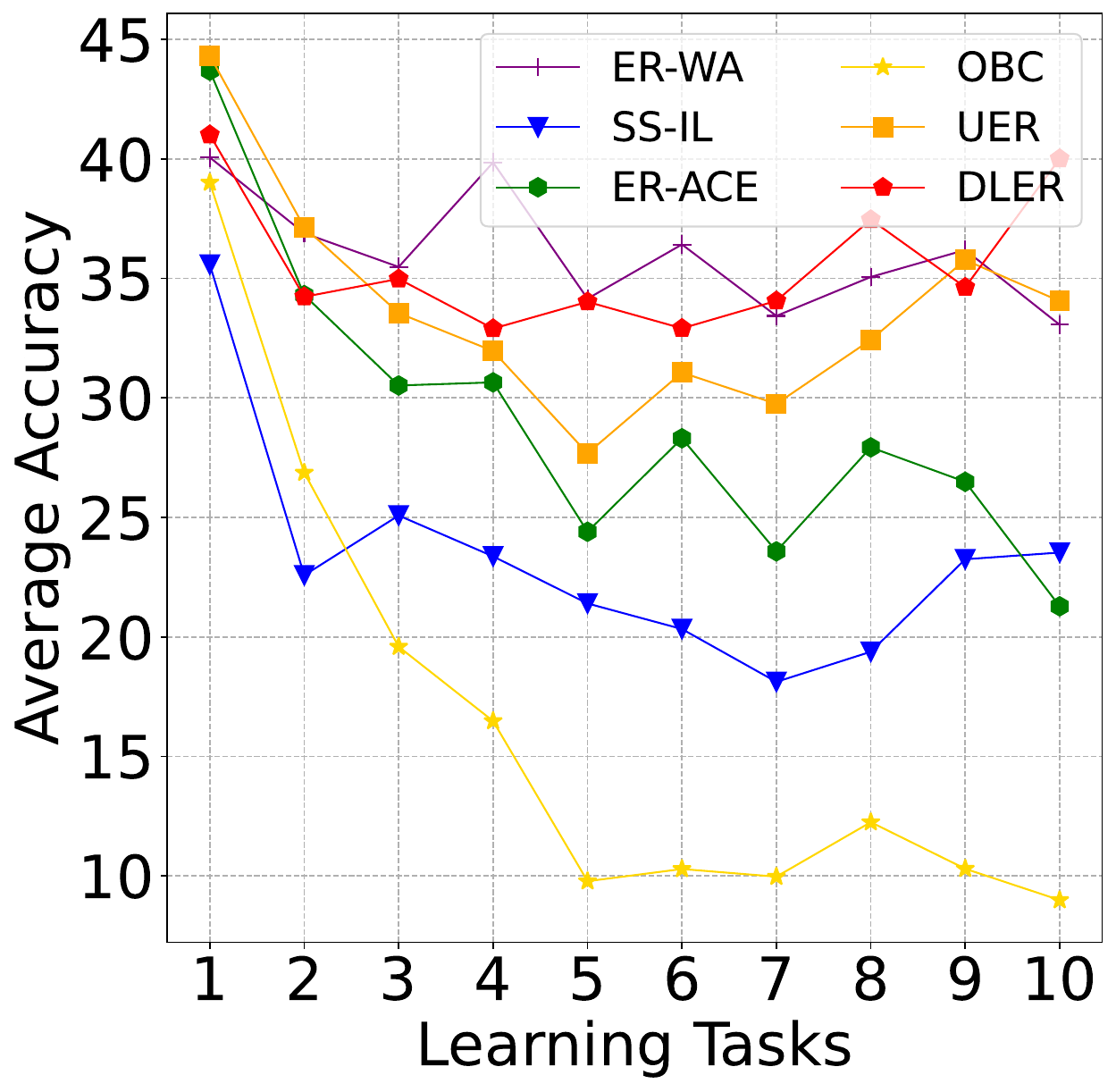}
    \end{minipage}
}
\subfigure[Average of Previous and Current Classes]{
    \begin{minipage}[t]{0.31\linewidth}
        \centering
        \includegraphics[scale=0.26]{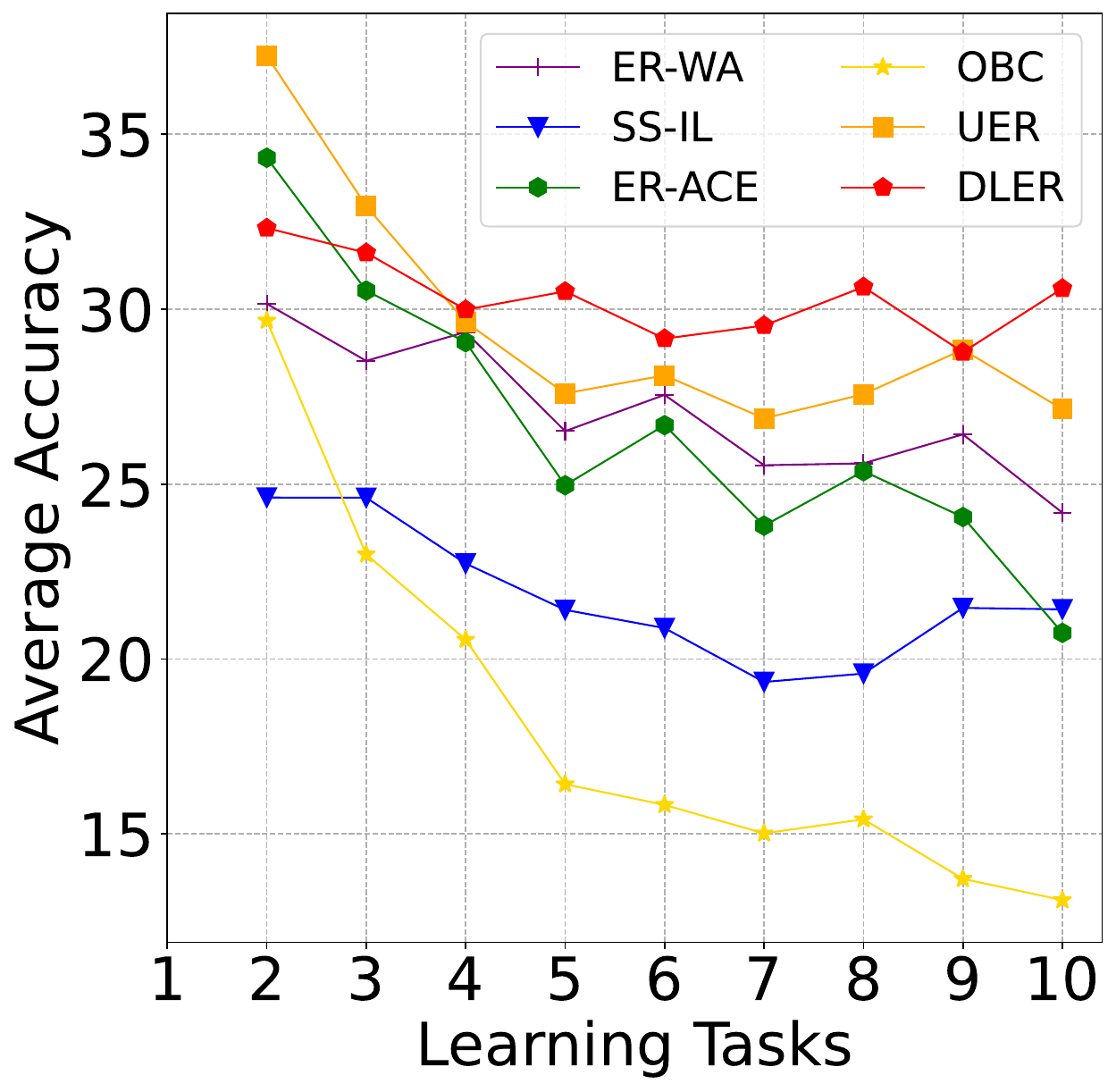}
    \end{minipage}
}
\centering
\vspace{-2ex}
\caption{Average accuracy rate of previous classes, current classes, and their average on Split MiniImageNet with a 1000 memory buffer size. }
\label{fig:balance}
\vspace{-3ex}
\end{figure*}

\begin{figure*}[t]
\centering
\subfigure[Split CIFAR100]{
    \begin{minipage}[t]{0.31\linewidth}
        \centering
        \includegraphics[scale=0.26]{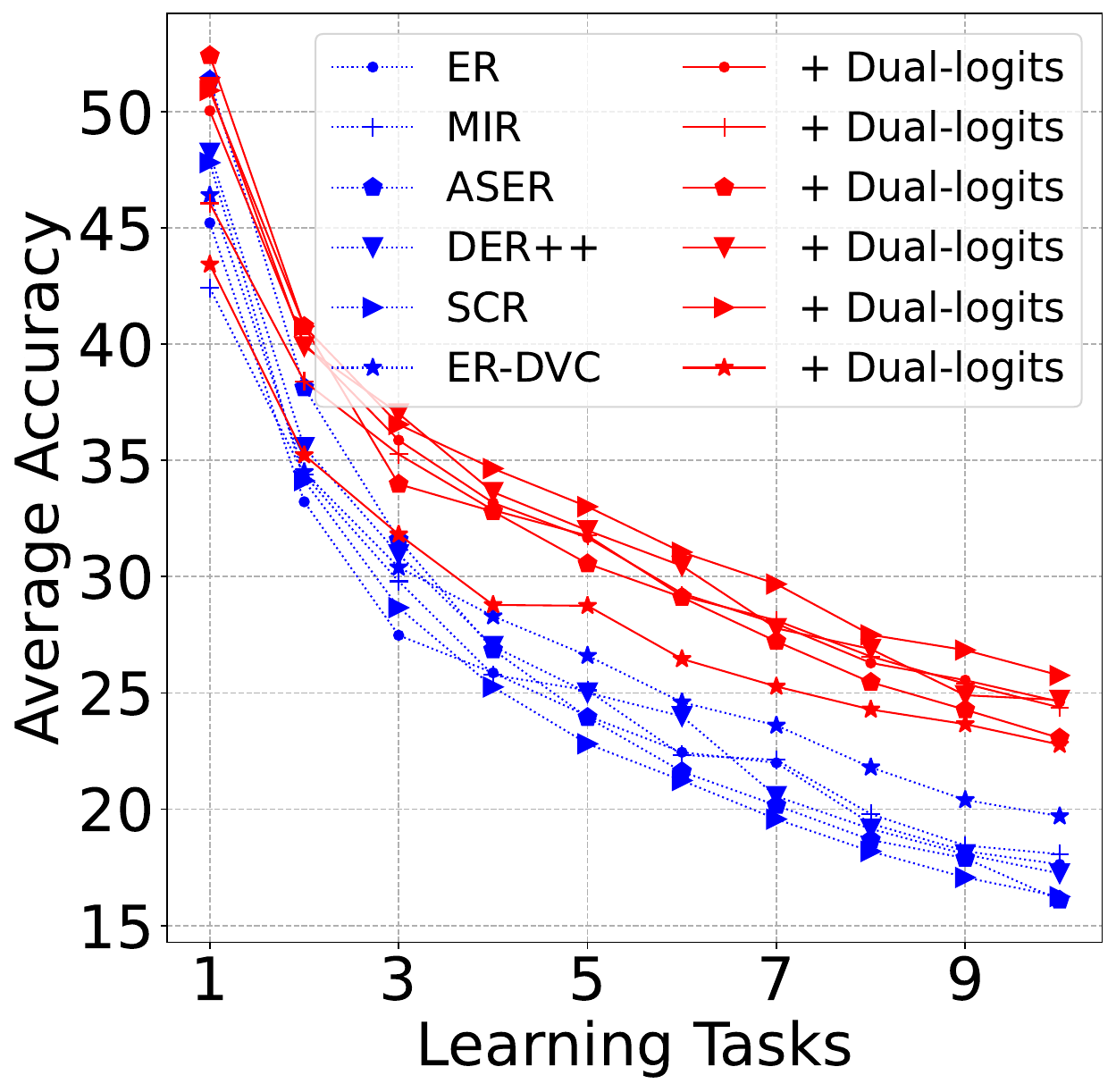}
    \end{minipage}
}
\subfigure[Split MiniImageNet]{
    \begin{minipage}[t]{0.31\linewidth}
        \centering
        \includegraphics[scale=0.26]{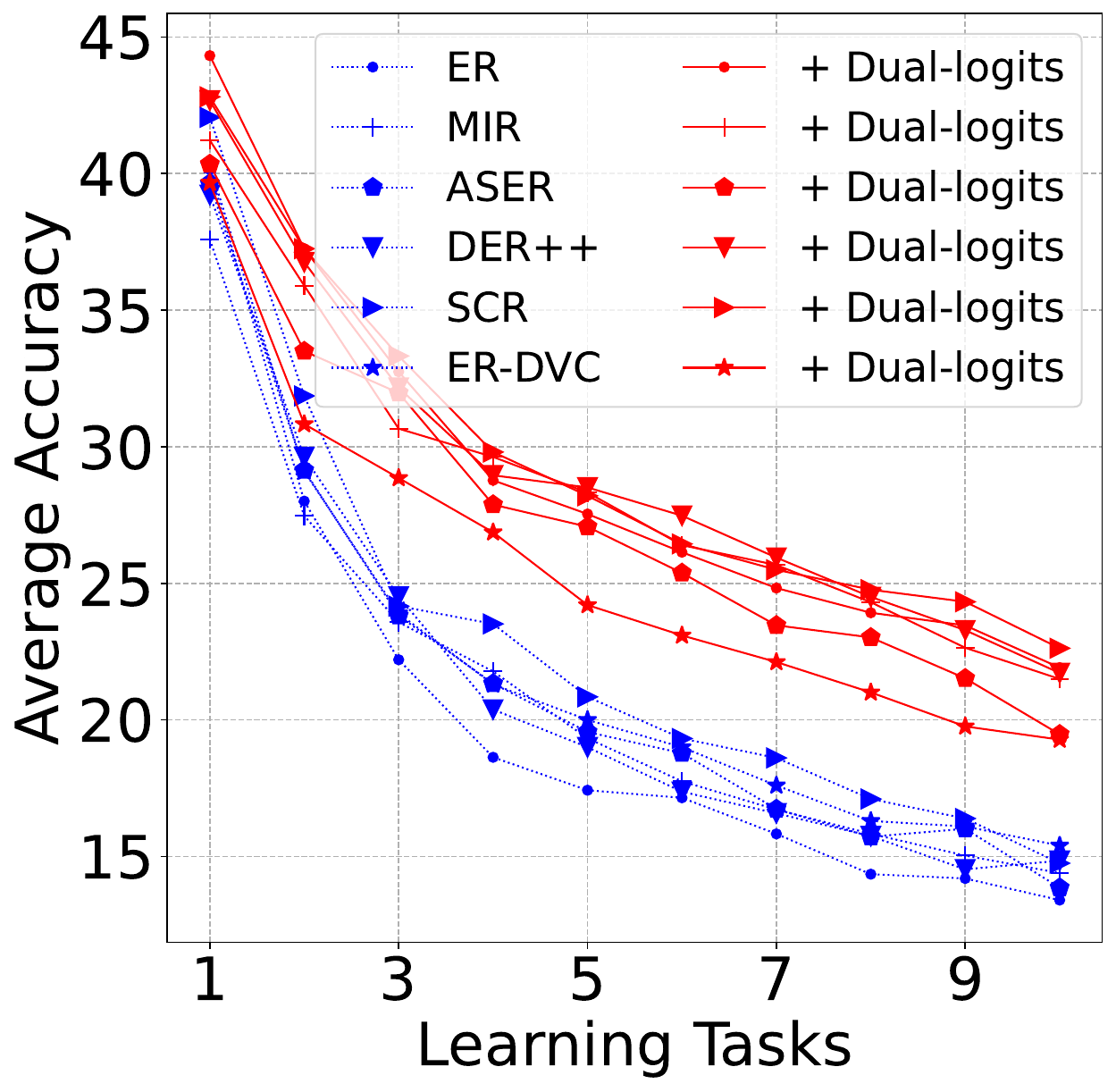}
    \end{minipage}
}
\subfigure[Split TinyImageNet]{
    \begin{minipage}[t]{0.31\linewidth}
        \centering
        \includegraphics[scale=0.26]{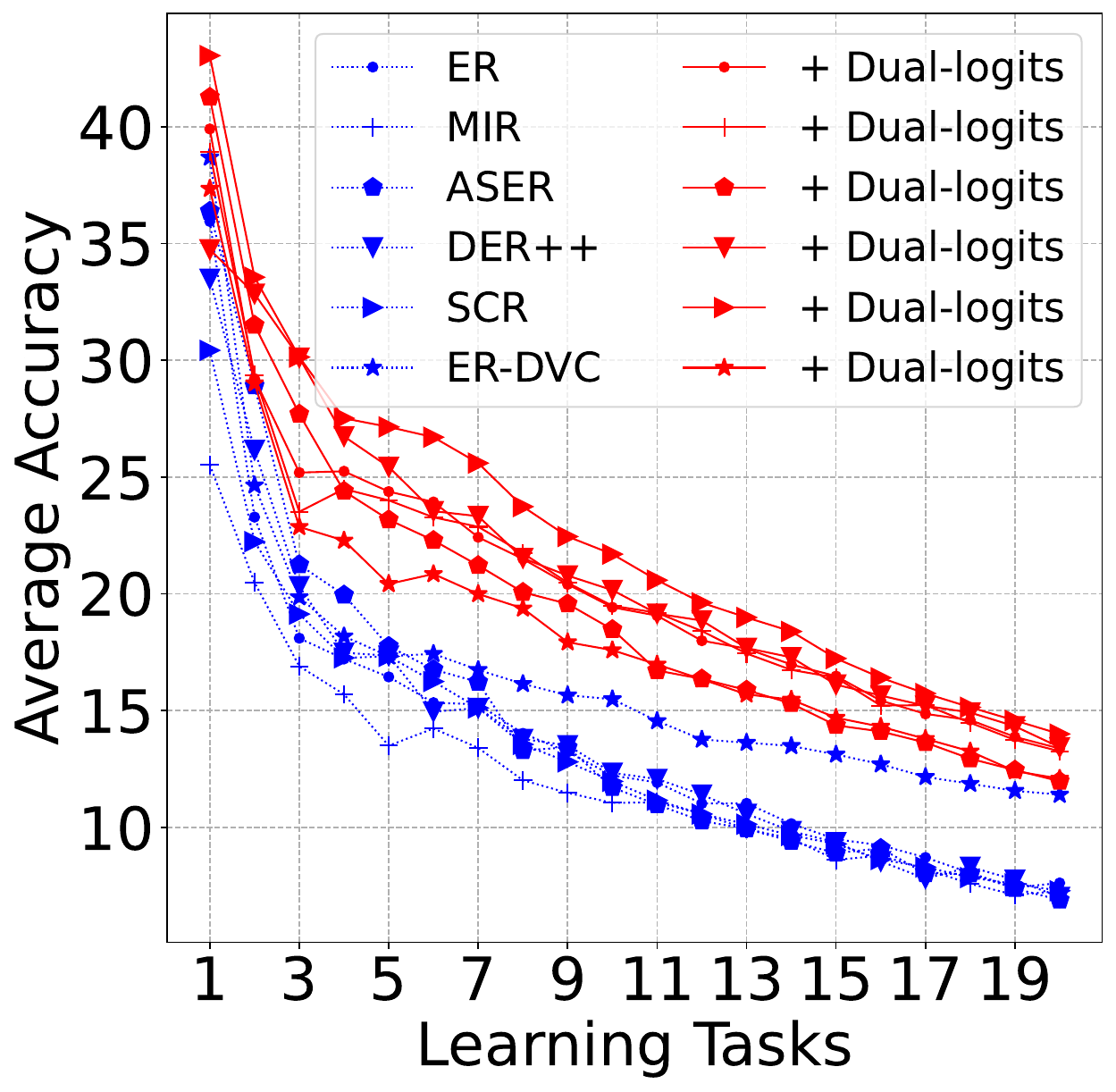}
    \end{minipage}
}
\centering
\vspace{-2ex}
\caption{Improvement of existing methods with dual-logits training on three datasets while the buffer size is 1000. }
\label{improvement}
\vspace{-3ex}
\end{figure*}

\begin{figure}[t]
\centering
\includegraphics[scale=0.20]{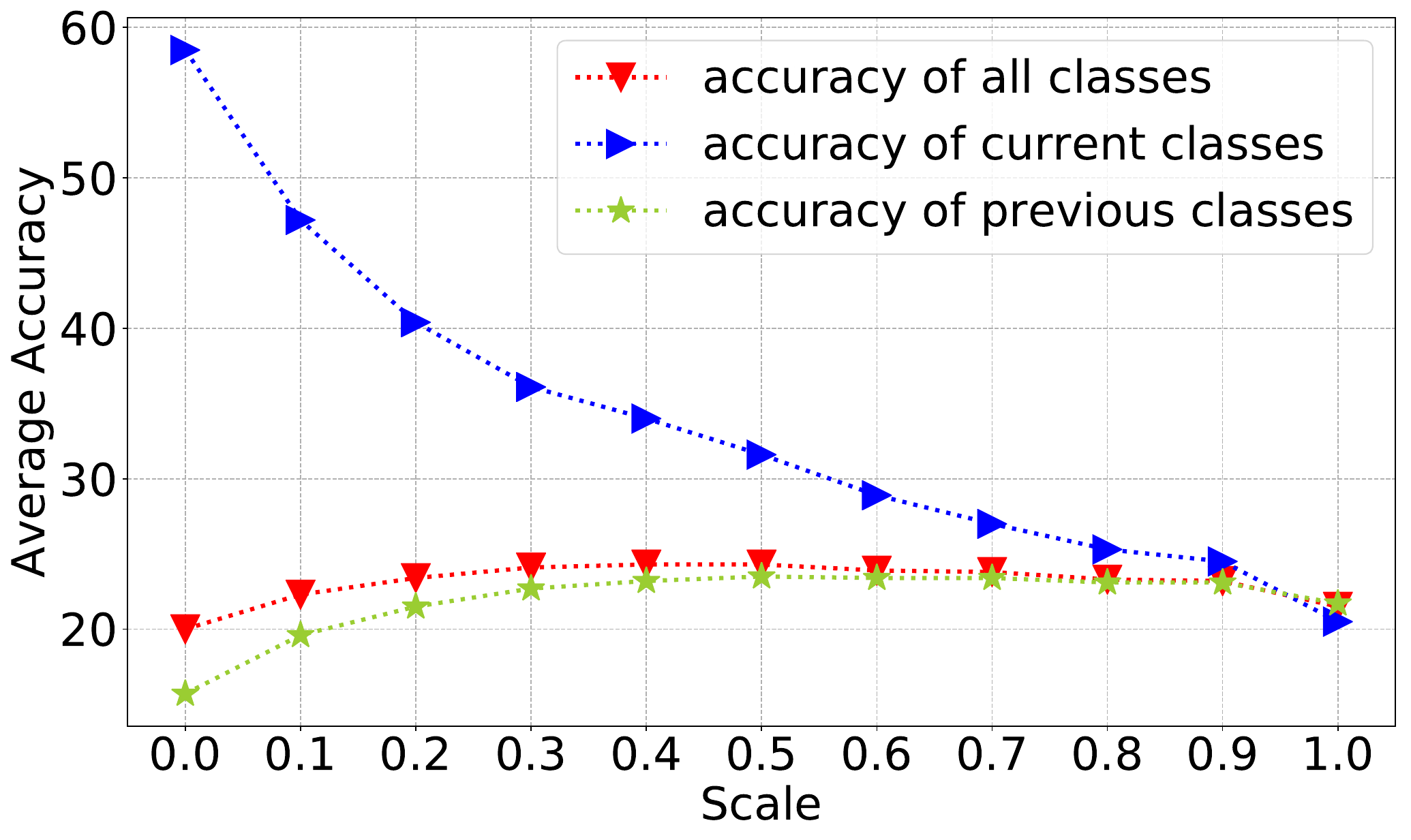}
\vspace{-2ex}
\caption{The performance of dual-logits training on Split CIFAR100 (buffer size=1000) with different values of $\alpha$.}
\label{fig:scale}%
\vspace{-2ex}
\end{figure}

\subsection{Ablation Study}
Table~\ref{tab:ablation_tinyimagenet} presents an ablation study on Split TinyImageNet with buffer sizes of 400, 1000, 2000, and 4000. Starting from ER, we progressively introduce dual-logit training, controllable $\alpha$, supervised contrastive loss (SC), balanced memory (BM), and advantage-based sample selection. Dual-logit training combines cosine supervision for current samples with additional dot-product supervision during replay, whose relative contribution is controlled by $\alpha$. SC promotes feature compactness, BM balances memory allocation, and advantage-based selection favors samples with a positive dot-product probability advantage for the correct class. Each successive configuration improves mean final accuracy across all four capacities, with DLER achieving the best results.

\subsubsection{Dual-Logit Training} Dual-logit training incorporates norm information through dot-product supervision to balance current learning and historical retention. Figure~\ref{fig:balance} compares accuracy on current classes, previous classes, and their equally weighted average on Split MiniImageNet. Dual-logit training, represented by UER, maintains competitive accuracy on both groups and a higher equally weighted average than these baselines throughout learning. This balance is controlled by $\alpha$, which adjusts the relative contributions of dot-product and cosine supervision during replay.

Meanwhile, dual-logit training offers \textbf{flexibility} in where the mechanism is applied. To examine this property, we add a linear layer to the feature extractor and apply the mechanism to this layer, the predictor, or both. Table~\ref{table:flexibility} reports the final accuracy on all classes, current classes, and previous classes. Applying the mechanism to either layer improves overall accuracy over ER, while applying it to both achieves the highest overall accuracy. Compared with the predictor, using both layers improves the performance of previous classes while maintaining competitive performance of current classes.

\begin{table}[t]
\centering
\setlength{\tabcolsep}{6pt}
\renewcommand{\arraystretch}{1.1}
\caption{Final accuracy (\%) on Split CIFAR100 with a
buffer of 1,000 samples after adding a linear layer to the
feature extractor. }
\vspace{-1ex}
\label{table:flexibility}
\begin{tabular}{@{}lccc@{}}
\toprule
\shortstack[l]{Method\\Setting} & \shortstack[l]{All\\Classes} & \shortstack[l]{Current\\Classes} & \shortstack[l]{Previous\\Classes} \\
\midrule
ER                      & $15.6 \pm 2.6$ & $50.5$ & $11.7$ \\
ER-ACE                  & $18.9 \pm 0.9$ & $17.3$ & $19.1$ \\
OBC                     & $18.3 \pm 0.7$ & $17.5$ & $18.4$ \\
Dual-logit training (feature) & $18.6 \pm 1.0$ & $30.6$ & $17.2$ \\
Dual-logit training (predictor)         & $19.6 \pm 1.5$ & $34.7$ & $18.0$ \\
Dual-logit training (both)              & $\mathbf{20.1 \pm 1.2}$
                                       & $31.8$ & $18.8$ \\
\bottomrule
\end{tabular}
\vspace{-4ex}
\end{table}

Additionally, dual-logits training can be used with existing methods. Figure~\ref{improvement} compares six ER-based baselines with and without dual-logit training on the three datasets. The blue curves represent the original baselines, while the red curves show their performance after integration. All six baselines achieve higher final accuracy, with improvements throughout most of the learning process. These results suggest that incorporating norm information during replay can complement existing methods for mitigating prediction bias.

\subsubsection{Controllable $\alpha$} To examine how $\alpha$ influences this balance, we conduct experiments on Split CIFAR100 with different values of $\alpha$ in Figure~\ref{fig:scale}. The accuracy on current classes generally decreases as $\alpha$ increases, whereas the accuracy on previous classes initially improves and then declines. Consequently, the overall accuracy first increases but saturates or decreases when $\alpha$ becomes too large. Consistently, Table~\ref{tab:ablation_tinyimagenet} shows that allowing $\alpha$ to be adjusted improves performance over fixing $\alpha=1$ across all four memory capacities. These results support the importance of balancing the two objectives and demonstrate the \textbf{adjustability} of our proposed method.

\begin{figure}[t]
\centering
\includegraphics[scale=0.35]{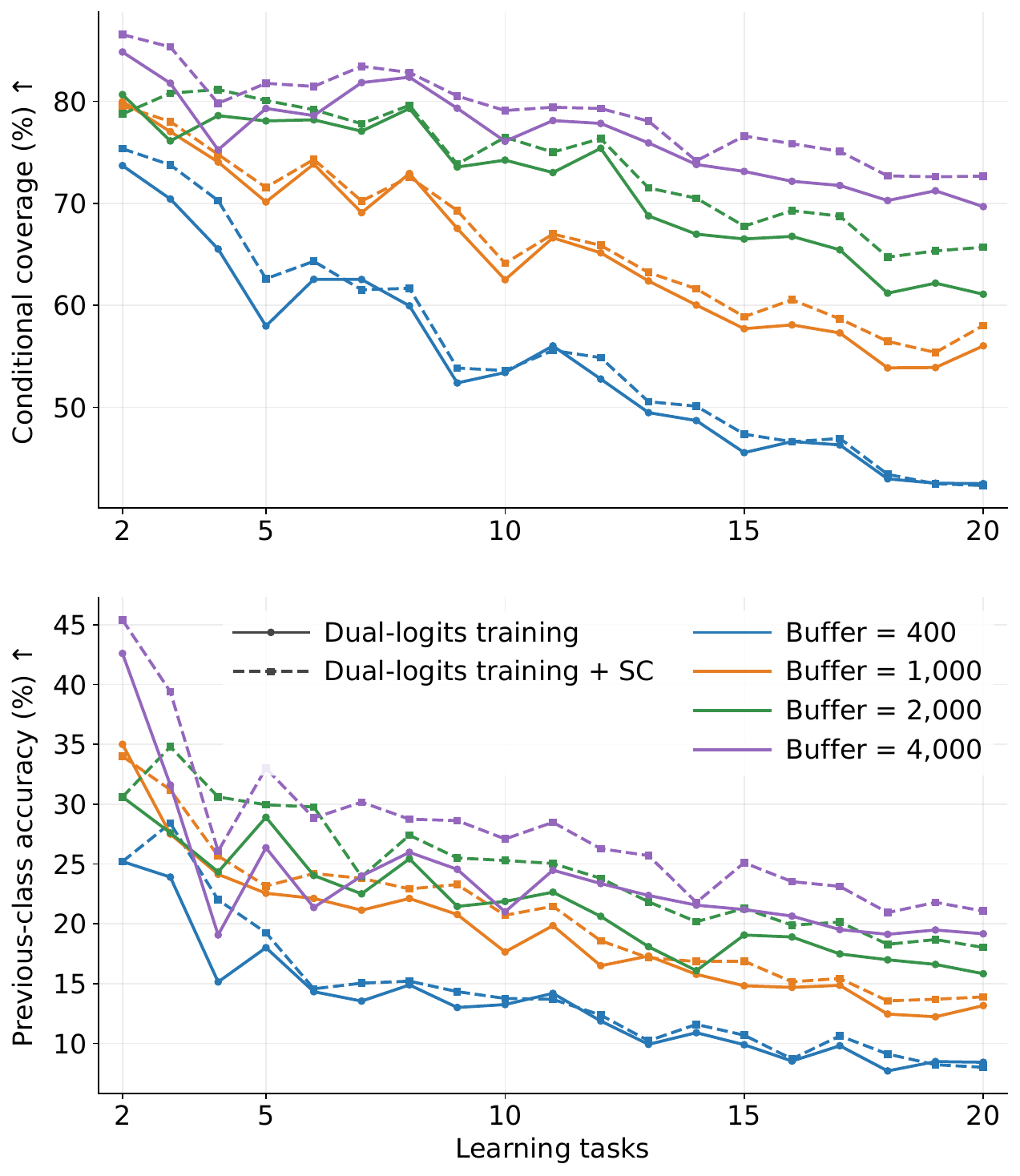}
\vspace{-2ex}
\caption{Conditional coverage and previous-class accuracy of dual-logits training and its enhancement with SC on Split TinyImageNet. }
\label{fig:sc_analysis}%
\vspace{-4ex}
\end{figure}

\subsubsection{Supervised contrastive objective}
SC is proposed to enhance the dual-logits training mechanism. To investigate how SC enhances dual-logit training, Figure~\ref{fig:analysis} compares training with and without SC on Split TinyImageNet using identical memory contents and test sample pairings. We measure conditional coverage and previous-class accuracy. Conditional coverage is the proportion of test samples from a previous class correctly distinguished from a given current class, conditioned on a stored reference from the same previous class being correctly distinguished from that current class. It measures whether separation achieved by stored samples extends to unseen samples of the same class. Previous-class accuracy evaluates previous-class test samples over all observed classes.

SC generally improves both metrics, with more consistent gains under larger buffers and improvements at every evaluated stage for the largest buffer. With the smallest buffer, gains fluctuate, and final performance is slightly lower despite higher stage averages. These trends suggest that SC helps replay supervision generalize beyond stored samples. By promoting within-class angular compactness and between-class separation, SC provides more consistent angular representations for both logits, potentially extending corrections learned through dot-product supervision to other samples of the same class.

\subsubsection{Balanced Memory component}

Reservoir sampling approximately preserves the class distribution of the data stream, yielding balanced memory in expectation for a balanced stream. However, score-based selection may favor certain classes, motivating BM to regulate memory allocation. To separate its contribution from advantage-based selection, Table~\ref{tab:ablation_tinyimagenet} includes dual-logit training with SC and BM as an intermediate configuration. Compared with SC and reservoir sampling, adding BM yields modest gains across all four memory capacities, suggesting that class balance alone does not explain the full improvement of DLER.

\subsubsection{The Advantage component}
Building on BM, the advantage component prioritizes samples for which dot-product prediction assigns a higher probability to the correct class than cosine prediction. As shown in Table~\ref{tab:ablation_tinyimagenet}, this component further improves mean accuracy across all memory capacities, with larger gains under smaller buffers. To investigate the characteristics associated with this score, we analyze the final DLER model trained on Split TinyImageNet with a buffer size of 400. We focus on the 96 classes for which DLER achieves higher accuracy than the configuration using dual-logit training with SC and CB. Within each class, we select the highest- and lowest-scoring samples using the final model and compare their spatial responses, feature statistics, and gradients. The results in Table~\ref{tab:dler_sample_analysis} characterize these selected classes.

\begin{figure*}[t]
    \centering
    \includegraphics[scale=0.54]{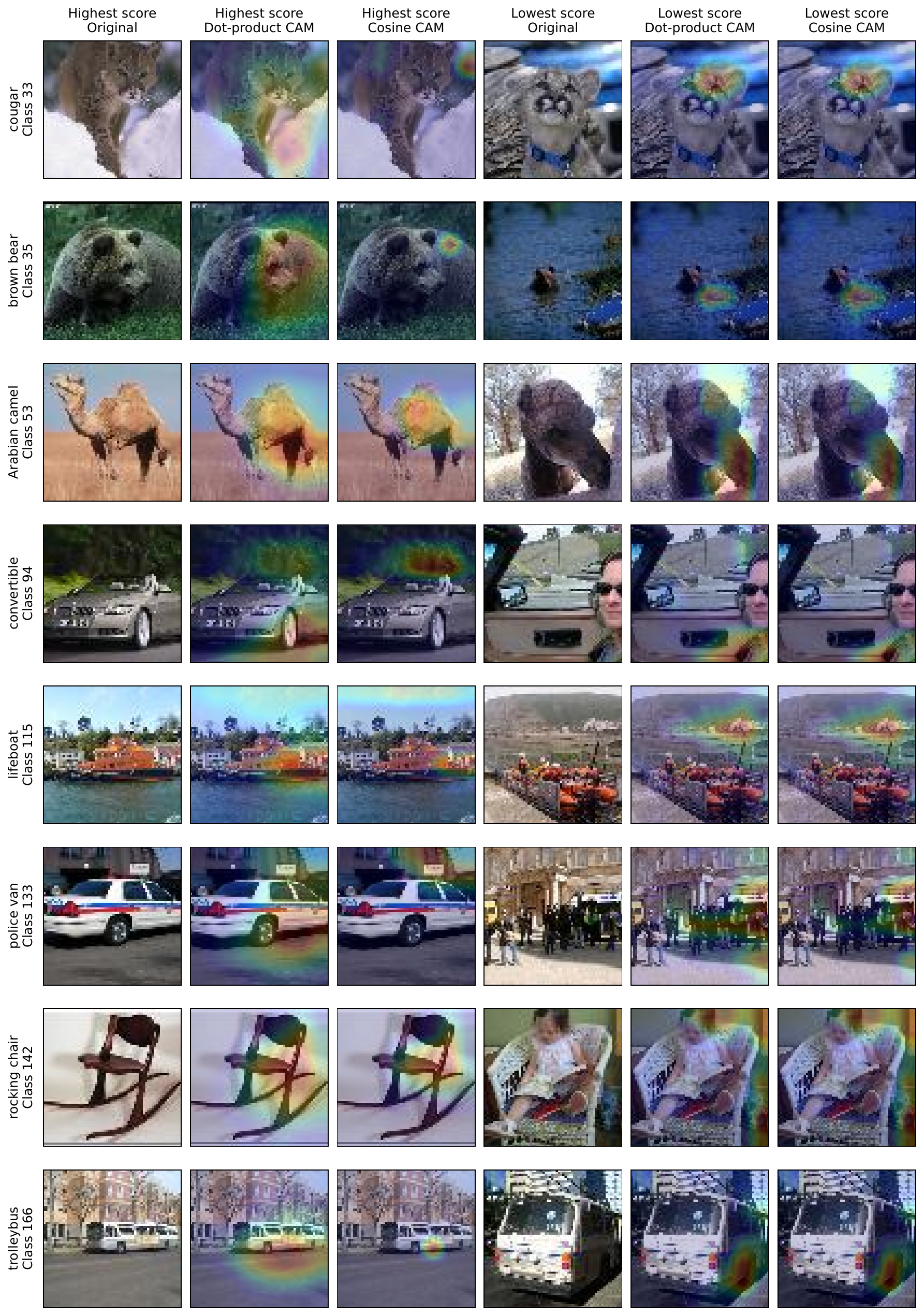}
    \vspace{-2ex}
    \caption{Examples of high-score and low-score samples and their class activation maps obtained using dot-product and cosine logits.}
    \label{fig:samples}
    \vspace{-2ex}
\end{figure*}

\begin{table}[t]
    \centering
    \caption{Comparison of the highest- and lowest-scoring training samples within the 96 selected classes on Split TinyImageNet.}
    \vspace{-1ex}
    \label{tab:dler_sample_analysis}
    \small
    \setlength{\tabcolsep}{6pt}
    \renewcommand{\arraystretch}{1.1}
    \begin{tabular}{@{}lrr@{}}
        \toprule
        Metric & High-score & Low-score \\
        \midrule
        Spatial JS divergence & 0.2256 & 0.0463 \\
        Feature norm & 15.7856 & 12.8475 \\
        Ground-truth cosine probability & 0.1173 & 0.0260 \\
        Gradient norm (all) & 7.3114 & 52.7980 \\
        Gradient norm (backbone) & 7.0487 & 51.4370 \\
        Gradient conflict rate & 41.5\% & 44.6\% \\
        \bottomrule
    \end{tabular}
    \vspace{-4ex}
\end{table}
We compare high- and low-score samples from three perspectives. 1) \textbf{Spatial responses.} Class activation maps (CAMs) are extracted from \texttt{layer4} using the ground-truth dot-product and cosine logits and normalized by their spatial sums. High-score samples exhibit larger spatial Jensen--Shannon (JS) divergence, indicating more distinct responses between the two logits, as illustrated in Figure~\ref{fig:samples}. 2) \textbf{Feature characteristics.} High-score samples have larger feature norms and higher ground-truth cosine probabilities, suggesting that their dot-product advantage coexists with stronger feature magnitudes and useful angular evidence rather than simply weak cosine predictions. 3) \textbf{Gradient behavior.} Under combined cosine and dot-product supervision, high-score samples exhibit substantially smaller overall and backbone gradient norms and a slightly lower gradient conflict rate, measured as the proportion of sample pairs with negative gradient cosine similarity. This indicates milder updates and less conflict among evaluated pairs. Overall, advantage-based selection identifies samples with stronger feature magnitudes, greater cosine support, and additional dot-product probability advantages, potentially supporting more effective replay under limited memory.

\begin{table}[t]
\centering
\caption{Theoretical comparison of training computation, model parameters, and inference computation. }
\label{tab:complexity}
\vspace{-1ex}

\begingroup
\footnotesize
\setlength{\tabcolsep}{3pt}
\renewcommand{\arraystretch}{1.15}

\begin{adjustbox}{max width=\columnwidth}
\begin{tabular}{@{}lccc@{}}
\toprule
Method
& \shortstack{Training\\per update}
& \shortstack{Model\\parameters}
& \shortstack{Inference\\per sample} \\
\midrule

ER
& $O(BQ)$
& $P$
& $O(F_{\mathrm{pred}})$ \\

ER-ACE
& $O(BQ)$
& $P$
& $O(F_{\mathrm{pred}})$ \\

SS-IL
& $O(BQ+BF_{\mathrm{pred}})$
& $P+P_{\mathrm{old}}$
& $O(F_{\mathrm{pred}})$ \\

UER
& $O(BQ)$
& $P$
& $O(F_{\mathrm{pred}})$ \\

\midrule
\textbf{DLER}
& 
$
O\bigl(BQ+B^2d+MF_{\mathrm{pred}}+T_{\mathrm{sel}}\bigr)
$
& $P$
& $O(F_{\mathrm{pred}})$ \\

\bottomrule
\end{tabular}
\end{adjustbox}
\endgroup
\vspace{-4ex}
\end{table}

\subsection{Complexity Analysis}
Table~\ref{tab:complexity} compares computational and parameter storage requirements. Here, $B$, $M$, and $d$ denote the combined training batch size, memory capacity, and feature dimension; $Q$ and $F_{\mathrm{pred}}$ denote per-sample training and prediction costs. $P$ and $P_{\mathrm{old}}$ are the standard and teacher model parameter counts, and $T_{\mathrm{sel}}$ is the memory selection cost. ER, ER-ACE, and UER share the same asymptotic training complexity, while SS-IL requires additional teacher predictions and storage. DLER retains UER's architecture without additional parameters.

DLER adds $O(B^2d)$ contrastive computation, $O(MF_{\mathrm{pred}})$ memory scoring, and sample selection. Scoring scales linearly with the fixed buffer capacity and can be batched without retaining backpropagation graphs. All methods require $O(MS)$ replay storage, where $S$ is the storage per sample; DLER adds $O(M+C)$ scores and metadata for $C$ observed classes. These additional operations occur only during training. DLER thus improves replay selection while preserving the model size and inference cost of ER and UER, supporting its \textbf{Real-time ability} through direct prediction without additional inference modules or post-processing.

\section{Conclusion}
\label{sec:conclusion}
In this paper, we propose DLER to alleviate catastrophic forgetting in OCL by addressing prediction bias. DLER learns current samples using cosine logits and incorporates additional dot-product supervision during replay. To further strengthen the dual-logit mechanism, an auxiliary supervised contrastive objective promotes angular compactness, supporting the generalization of corrections beyond limited replay samples. We also extend dual-logits to memory management through advantage-based scoring, aligning sample retention with the dual-logit training. Extensive experiments on three datasets demonstrate the effectiveness of DLER, particularly under limited replay memory.

\bibliographystyle{IEEEtran}

\section{Appendix}
\label{sec:appendix}

\subsection{Analysis of Catastrophic Forgetting}
There is a competition between samples of each class in the process of gradient propagation. By the chain rule, the gradient of the predictor for a training sample $\bm{x}$ of class $y$ can be expressed as
\begin{equation}
    \label{eq:gradient_w}
    \frac{\partial L_{er}}{\partial \bm{w}_j}=\left\{
    \begin{aligned}
        (p_y^{dot}-1)\bm{h}&,&j= y \\
        p_j^{dot}\bm{h}&,&j\neq y
    \end{aligned}
    \right.,
\end{equation}
and
\begin{equation}
    \label{eq:gradient_b}
    \frac{\partial L_{er}}{\partial b_j}=\left\{
    \begin{aligned}
        p_y^{dot}-1&,&j= y \\
        p_j^{dot}&,&j\neq y
    \end{aligned}
    \right..
\end{equation}
With a positive learning rate $\eta$ and a non-negative feature vector $\bm{h}$, the gradient updates increase the parameters associated with the ground-truth class $y$: $\bm{w}_y\leftarrow\bm{w}_y-\eta(p_y-1)\bm{h}$ and $b_y\leftarrow b_y-\eta(p_y-1)$. Conversely, they decrease the parameters associated with other classes: $\bm{w}_j\leftarrow\bm{w}_j-\eta p_j\bm{h}$ and $b_j\leftarrow b_j-\eta p_j$, for $j\neq y$. These updates move $\bm{w}_y$ toward the direction of $\bm{h}$ and increase $b_y$. Thus, it increases the logits score of the $y$-th dimension as well as decreases the logits scores of other dimensions. And the process is the same when using cosine logits by
\begin{equation}
    \label{eq:gradient_w_cos}
    \frac{\partial L_{lucir}}{\partial \bm{w}_j}=\left\{
    \begin{aligned}
        (p_y^{cos}-1)\hat{\bm{h}}&,&j= y \\
        p_j^{cos}\hat{\bm{h}}&,&j\neq y
    \end{aligned}
    \right.,
\end{equation}
where
\begin{equation}
    \label{eq:h}
	\hat{\bm{h}}=\frac{\frac{\bm{h}}{\Vert\bm{h}\Vert}-cos(\bm{w},\bm{h})\frac{\bm{w}}{\Vert\bm{w}\Vert}}{\Vert\bm{w}\Vert}.
\end{equation}

The prediction bias is a phenomenon of CF, mainly caused by the unbalanced competition of gradient propagation. Generally, the number of samples belonging to previous classes in the fixed buffer will decrease as the learning process continues. The number of samples belonging to current classes remains unchanged. If the model incrementally learns by Equation~\eqref{eq:erloss} or Equation~\eqref{eq:lucirloss}, the associated parameters of current classes receive more positive changes, while the others obtain more negative changes. This means the gradients of previous classes lose their competitiveness. Besides, the logits of current classes tend to be larger than the previous ones. Hence, the model is more likely to classify most samples to the classes of current samples, reducing the performance of previous classes.

\begin{table*}[t]
\centering
\caption{Comparison of logit configurations for training and inference on Split CIFAR100 with a memory buffer of 5,000 samples. $A_{\mathrm{all}}$, $A_p$, and $A_c$ denote the final accuracies (\%) on all, previous, and current classes, respectively. $\overline{\|\bm{w}\|}$ denotes the mean class-weight vector norm; $\mu_W$ and $\sigma_W$ denote the mean and standard deviation of weight entries; and $\mu_b$ denotes the mean bias. Subscripts $p$ and $c$ indicate previous and current classes. Dot and Cos denote dot-product and cosine logits, respectively.}
\label{table:analysis}
\begingroup
\footnotesize
\setlength{\tabcolsep}{3.0pt}
\renewcommand{\arraystretch}{1.15}
\begin{tabular*}{\textwidth}
{@{\extracolsep{\fill}}lccc ccc cc cc cc@{}}
\toprule
& \multicolumn{3}{c}{Logit configuration}
& \multicolumn{3}{c}{Accuracy (\%)}
& \multicolumn{2}{c}{Weight norm}
& \multicolumn{2}{c}{Weight statistics}
& \multicolumn{2}{c}{Bias} \\
\cmidrule(lr){2-4}
\cmidrule(lr){5-7}
\cmidrule(lr){8-9}
\cmidrule(lr){10-11}
\cmidrule(l){12-13}
Configuration
& Learning & Replaying & Testing
& $A_{\mathrm{all}}$ & $A_p$ & $A_c$
& $\overline{\|\bm{w}\|}_p$
& $\overline{\|\bm{w}\|}_c$
& $\mu_{W_p}\pm\sigma_{W_p}$
& $\mu_{W_c}\pm\sigma_{W_c}$
& $\mu_{b_p}$ & $\mu_{b_c}$ \\
\midrule
1 (ER)
& Dot & Dot & Dot
& 27.3 & 17.7 & 36.9
& 2.08 & 1.86
& $0.0044 \pm 0.0274$ & $-0.0038 \pm 0.0221$
& $-0.22$ & $0.22$ \\
2
& Dot & Dot & Cos
& 24.1 & 12.7 & 35.6
& 2.08 & 1.86
& $0.0044 \pm 0.0274$ & $-0.0038 \pm 0.0221$
& $-0.22$ & $0.22$ \\
\addlinespace[2pt]
3 (LUCIR)
& Cos & Cos & Cos
& 26.9 & 14.9 & 38.9
& 2.00 & 1.64
& $0.0398 \pm 0.0235$ & $0.0471 \pm 0.0146$
& $-0.00$ & $0.00$ \\
4
& Cos & Cos & Dot
& 28.2 & 28.8 & 27.6
& 2.00 & 1.64
& $0.0398 \pm 0.0235$ & $0.0471 \pm 0.0146$
& $-0.00$ & $0.00$ \\
\addlinespace[2pt]
5
& Dot & Cos & Dot
& 19.6 & 0.0 & 39.1
& 1.37 & 2.09
& $0.0197 \pm 0.0115$ & $0.0219 \pm 0.0270$
& $-1.03$ & $1.03$ \\
6
& Dot & Cos & Cos
& 27.3 & 25.3 & 29.4
& 1.37 & 2.09
& $0.0197 \pm 0.0115$ & $0.0219 \pm 0.0270$
& $-1.03$ & $1.03$ \\
\addlinespace[2pt]
7
& Cos & Dot & Cos
& 21.3 & 0.5 & 42.2
& 2.52 & 1.51
& $0.0508 \pm 0.0371$ & $0.0565 \pm 0.0112$
& $0.20$ & $-0.20$ \\
8 (UER)
& Cos & Dot & Dot
& \textbf{29.5} & 30.9 & 28.1
& 2.52 & 1.51
& $0.0508 \pm 0.0371$ & $0.0565 \pm 0.0112$
& $0.20$ & $-0.20$ \\
\bottomrule
\end{tabular*}
\endgroup
\end{table*}

\subsection{Analysis of Biased Predictor}
Imbalanced gradient mainly affects the predictor through the angle and norm factors. The norm factor accounts for intra-class variation, and the angle factor accounts for inter-class variation~\cite{liu2018decoupled}. To study their changes in OCL, we conduct experiments on CIFAR100 with two incremental stages and 50 classes per stage. We set two general ways to calculate logits in the experiment: dot-product logits (both factors) and cosine logits (only angle factor). In Table~\ref{table:analysis}, we take different ways to calculate logits and record their performance.

Experiment results show that the characteristics of these two factors differ from each other. On the one hand, the bias issue mainly occurs in the angle factor, which efficiently learns novel knowledge. Comparing index 1 and 2, replacing dot-product logits with cosine logits in the testing phase leads to worse final accuracy, particularly for previous classes. Meanwhile, the method in index 3 performs better for current classes but worse for previous classes compared to the one in index 1. On the other hand, the norm factor, which has been abandoned in existing studies~\cite{hou2019learning}, benefits previous classes. As seen in indices 3 and 4, switching from cosine to dot-product logits during testing improves overall accuracy, with a substantial gain on previous classes.

With this observation, the two factors can be utilized in different occasions. The angle factor is effective in adapting novel knowledge with inter-class variation. Consequently, cosine logits that depend on the angle factor are suitable for learning current samples. And dot-product logits can be used for replaying previous samples since they can leverage the norm factor to enhance historical knowledge with intra-class variation. Motivated by these findings, we propose a method that learns current samples by cosine logits, replays previous samples, and predicts testing samples by dot-product logits.

\begin{figure*}[t]
\centering
\subfigure[The accumulate changes of parameters.]{
    \begin{minipage}[t]{0.3\linewidth}
        \centering
        \includegraphics[scale=0.15]{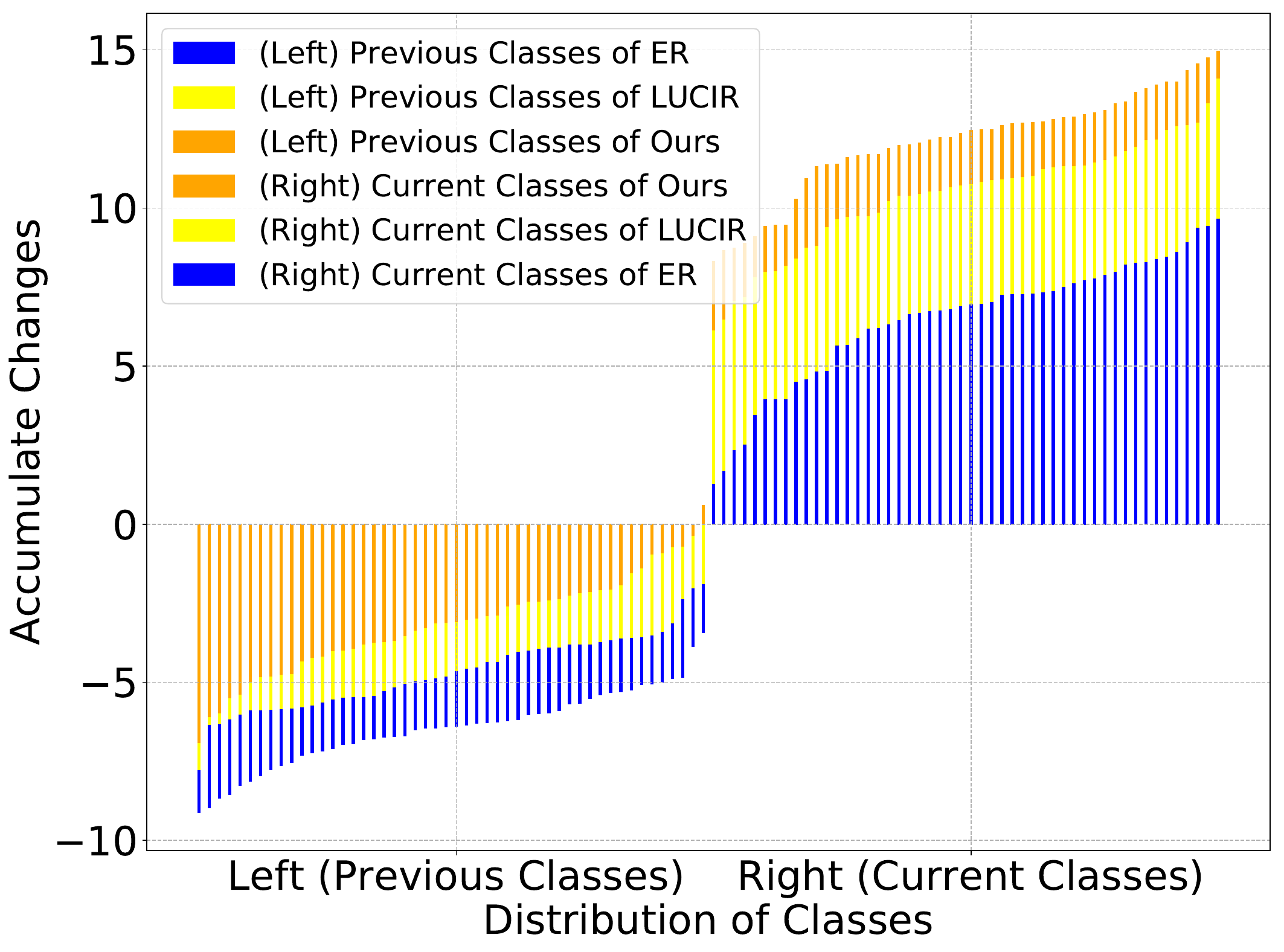}
    \end{minipage}
}
\hspace{.10in}
\subfigure[The average norm value of all classes.]{
    \begin{minipage}[t]{0.3\linewidth}
        \centering
        \includegraphics[scale=0.15]{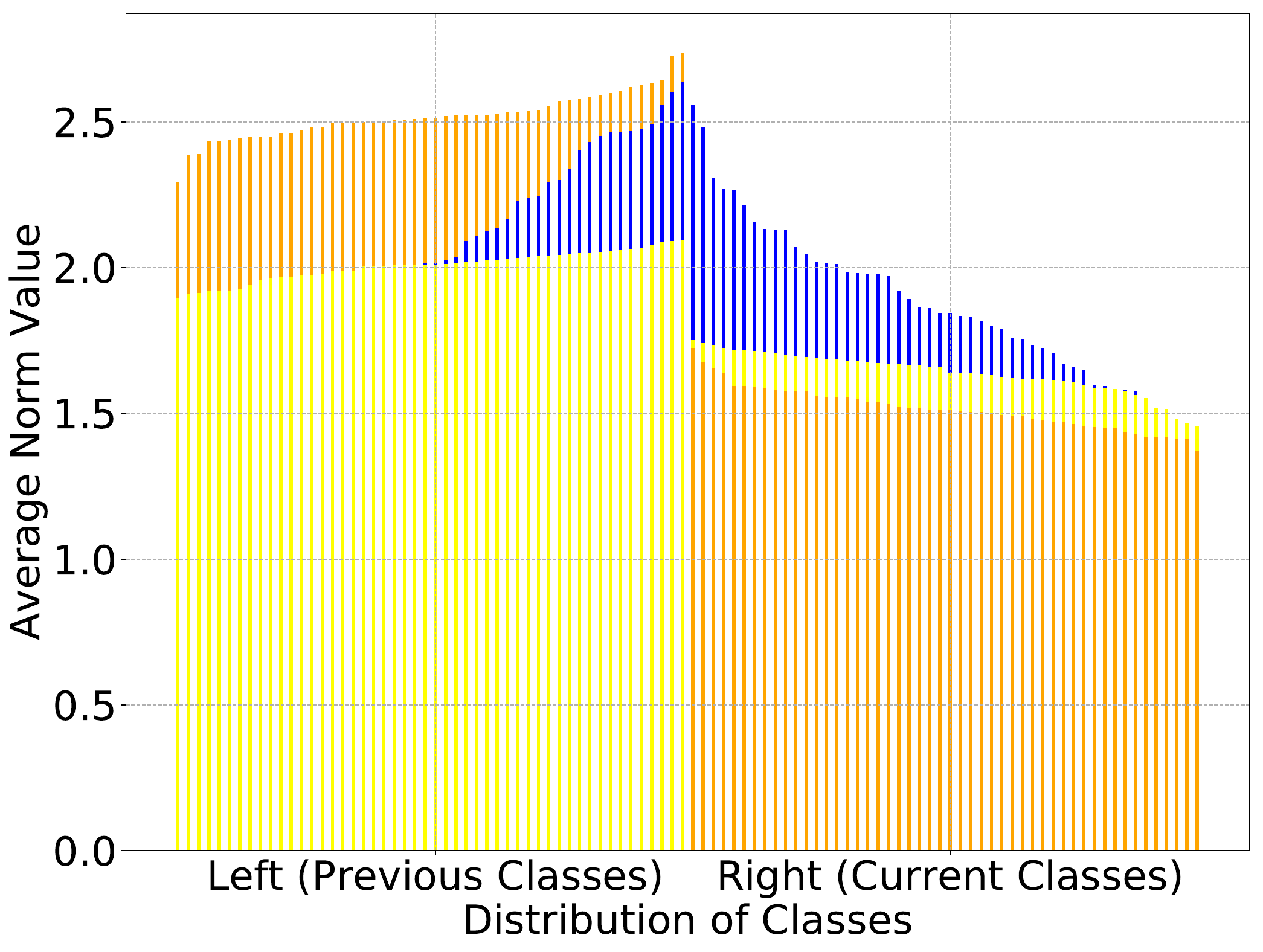}
    \end{minipage}
}
\hspace{.10in}
\subfigure[The categorical probability of testing samples.]{
    \begin{minipage}[t]{0.3\linewidth}
        \centering
        \includegraphics[scale=0.15]{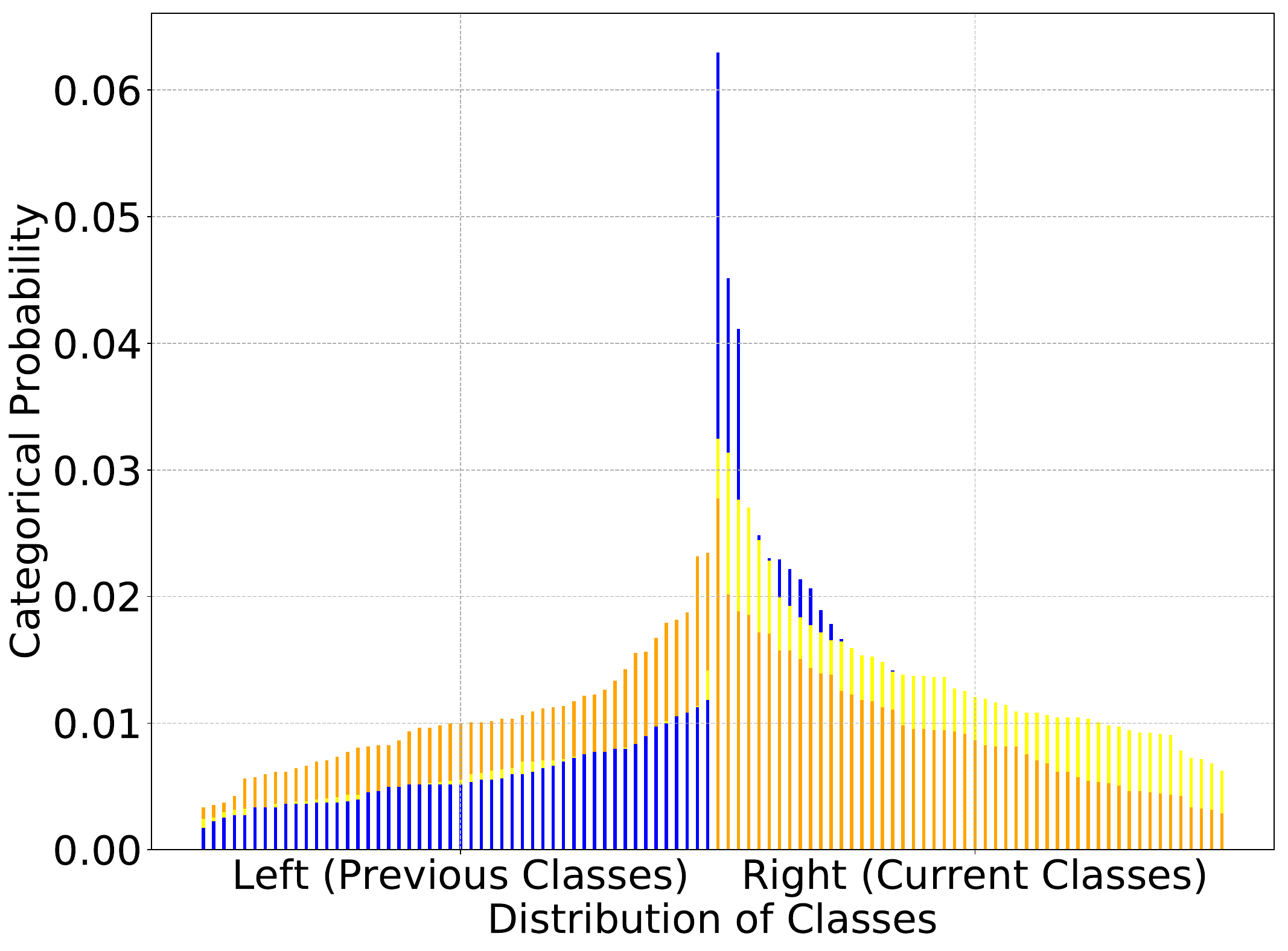}
    \end{minipage}
}
\centering
\vspace{-1ex}
\caption{The analysis results of different ways during the learning process on Split CIFAR100 when the memory buffer size is 5000. }
\label{fig:analysis}
\vspace{-1ex}
\end{figure*}

\subsection{Analysis of Motivation}
In addition to the results in Table~\ref{table:analysis}, we present more detailed results for ER (blue), LUCIR (yellow) and our method (orange) in Figure~\ref{fig:analysis}. We calculate the difference between the parameters of the predictor before and after learning the current classes, taking it as the accumulated changes in parameters as shown in Figure~\ref{fig:analysis} (a). Meanwhile, we report the average norm value of the weight vector for all classes in Figure~\ref{fig:analysis} (b). Moreover, we calculate the average categorical probability of all testing samples by the latest model, illustrated in Figure~\ref{fig:analysis} (c). Further analysis and experimental results verify the feasibility of our proposed method.

Firstly, learning current samples using cosine logits can save the novel knowledge into the angle factor. In indices 7 and 8, the performance on previous classes is almost zero when testing with cosine logits. The reason behind this is that the angular similarity is almost the determining factor for gradients in Equation (\ref{eq:gradient_w_cos}). For one thing, all vectors in Equation (\ref{eq:gradient_w_cos}) are normalized, and this greatly enhances the influence of the angular similarity. For another thing, the change of $\bm{w}$ is constrained by $\hat{\bm{h}}$, as large $\Vert\bm{w}\Vert$ can decrease gradients so much that the weights pass is not able to update effectively. Moreover, the dot-product logits is associated with a value range $[-\infty,+\infty]$, while the value range of cosine logits is $[-1,1]$. This causes the cosine logits to produce smaller changes for previous classes and larger changes for current classes (as seen in the blue and orange parts in Figure~\ref{fig:analysis} (a)). It also raises the norm factor of previous classes and reduces the norm factor of current classes (as observed in the blue and orange parts in Figure~\ref{fig:analysis} (b)), which helps learn novel knowledge by the angle factor.

Secondly, replaying previous samples with dot-product logits can enhance historical knowledge through the norm factor. Comparing index 7 and 8, using dot-product logits in the testing process results in the best performance for previous classes, even surpassing the performance of current classes. Different from cosine logits, the gradient in Equation (\ref{eq:gradient_w}) depends on both two factors. In the meantime, the gradient in Equation (\ref{eq:gradient_w}) is only associated with previous samples in the buffer. Since the angle factor is easily influenced by an unbalanced gradient, this process tends to pay more attention to the norm factor. As a result, replaying by dot-product logits not only decreases the changes for previous classes but also increases the changes for current classes (as seen in the yellow and orange parts in Figure~\ref{fig:analysis} (a)). In the meantime, it raises the norm factor of previous classes as well as reduces the norm factor of current classes (indicated by the yellow and orange parts in Figure~\ref{fig:analysis} (a)). It causes  that $\Vert\bm{W}_p\Vert$ is larger than $\Vert\bm{W}_c\Vert$, and Mean($b_c$) is smaller than Mean($b_p$). Although the angle factor suffers from $cos(\bm{w}_c,\bm{h})>cos(\bm{w}_p,\bm{h})$, it can be leveraged by $\Vert\bm{w}_p\Vert\Vert\bm{h}\Vert+b_p>\Vert\bm{w}_c\Vert\Vert\bm{h}\Vert+b_c$, as we described in Figure~\ref{fig:problem} (d).

As shown in Table~\ref{table:analysis}, our proposed method achieves the best performance among all combinations of different logits. Based on the combination of different logits, our method quickly learns novel knowledge and effectively retains historical knowledge. As shown in Figure~\ref{fig:analysis} (c), the posterior probability distribution of our method is relatively balanced, whereas those of ER and LUCIR tend to favor the current classes. Therefore, our motivation for learning current samples
by cosine logits while replaying previous samples by dot-product logits is reasonable and practicable.

\newpage

 




\vfill

\end{document}